\documentclass[final]{cvpr}

\usepackage{times}
\usepackage{epsfig}
\usepackage{graphicx}
\usepackage{amsmath}
\usepackage{amssymb}
\usepackage{algorithm}
\usepackage{algpseudocode}
\usepackage{float}
\usepackage{color}
\usepackage{colortbl}
\usepackage{breqn}
\usepackage{rotating}
\usepackage{enumitem}
\usepackage{makecell}
\usepackage{lineno}
\usepackage{tabularx}
\usepackage{xcolor}
\usepackage{multirow}
\usepackage{hhline}
\usepackage{amsfonts}
\usepackage{empheq}
\usepackage{float}
\usepackage{framed, color}
\usepackage{xcolor}
\usepackage{graphics}
\usepackage{graphicx}
\usepackage[]{graphicx} 
\usepackage{epsfig} 
\usepackage{color}
\usepackage{filecontents}
\usepackage{subfigure}
\usepackage{multirow}
\usepackage{filecontents}
\usepackage{verbatim} 
\usepackage{tabularx}
\usepackage{lineno}
\usepackage{setspace}
\usepackage{textcomp,booktabs}
\usepackage{caption}
\usepackage{stmaryrd}
\usepackage[mathscr]{eucal}
\usepackage{lineno}
\usepackage{subfigure}
\usepackage{bbm}
\usepackage[export]{adjustbox}

\def\D{{\bf D}}

\def\K{{\bf K}}

\def\R{{\bf R}}

\def\p{{\bf p}}
\def\q{{\bf q}}

\def\z{{\bf z}}

\def\M{{\bf M}}

\def\n{{\bf n}}
\def\U{{\bf U}}

\def\0{{\bf 0}}
\def\1{{\bf 1}}

\definecolor{red}{rgb}{0.95,0.4,0.4}
\definecolor{purered}{rgb}{1,0,0}
\definecolor{blue}{rgb}{0.4,0.4,0.95}
\definecolor{darkblue}{rgb}{0,0,0.8}
\definecolor{darkred}{rgb}{0.8,0,0}
\definecolor{darkgreen}{rgb}{0,0.5,0}
\definecolor{grey}{rgb}{0.6,0.6,0.6}
\definecolor{col1}{RGB}{232, 161, 148}
\definecolor{col2}{RGB}{148, 187, 232}
\definecolor{lightgrey}{rgb}{0.85,0.85,0.85}
\definecolor{alizarin}{rgb}{0.82, 0.1, 0.26}
\definecolor{bostonuniversityred}{rgb}{0.8, 0.0, 0.0}
\definecolor{aurometalsaurus}{rgb}{0.43, 0.5, 0.5}
\definecolor{shadegray}{rgb}{0.36, 0.36, 0.36}

\newcommand\boldred[1]{\textcolor{bostonuniversityred}{\bf #1}}

\graphicspath{{./figure/}}   

\usepackage[pagebackref=true,breaklinks=true,colorlinks,bookmarks=false]{hyperref}

\def\cvprPaperID{2219} 
\def\confYear{CVPR 2021}

\begin{document}

\title{\emph{Camera Pose Matters:} Improving Depth Prediction\\ by Mitigating Pose Distribution Bias}

\author{%
Yunhan Zhao$^1$ \quad Shu Kong$^2$ \quad Charless Fowlkes$^1$ \\
$^1$UC Irvine  \qquad $^2$Carnegie Mellon University \\
{\tt\small \{yunhaz5, fowlkes\}@ics.uci.edu } \ \ {\tt\small shuk@andrew.cmu.edu} \\ \\
\ [\href{https://www.ics.uci.edu/~yunhaz5/cvpr2021/cpp.html}{\boldred{Project Page}}] \quad
[\href{https://github.com/yzhao520/CPP}{\boldred{Github}}] \quad
[\href{https://www.ics.uci.edu/~yunhaz5/cvpr2021/02219-slides.pdf}{\boldred{Slides}}] 
\vspace{-5mm}
}

\maketitle

\begin{abstract}
Monocular depth predictors are typically trained on large-scale training sets which are naturally biased w.r.t the distribution of camera poses. As a result, trained predictors fail to make reliable depth predictions for testing examples captured under uncommon camera poses. To address this issue, we propose two novel techniques that exploit the camera pose during training and prediction. First, we introduce a simple perspective-aware data augmentation that synthesizes new training examples with more diverse views by perturbing the existing ones in a geometrically consistent manner. Second, we propose a conditional model that exploits the per-image camera pose as prior knowledge by encoding it as a part of the input. We show that jointly applying the two methods improves depth prediction on images captured under uncommon and even never-before-seen camera poses. We show that our methods improve performance when applied to a range of different predictor architectures. Lastly, we show that explicitly encoding the camera pose distribution improves the generalization performance of a synthetically trained depth predictor when evaluated on real images.
\end{abstract}

\section{Introduction}

Monocular depth prediction aims to estimate 3D scene geometry from a 2D image.
Despite being a largely underdetermined problem, convolutional neural network (CNN) based depth predictors trained on a sufficiently large-scale dataset are able to learn the joint statistics of scene geometry and appearance, and achieve impressive performance~\cite{eigen2014depth, eigen2015predicting, fu2018deep, guo2018learning, laina2016deeper, yin2019enforcing}.

\begin{figure}[t]
\centering
\includegraphics[width=\linewidth, trim=0mm 1mm 0mm 0mm, clip]{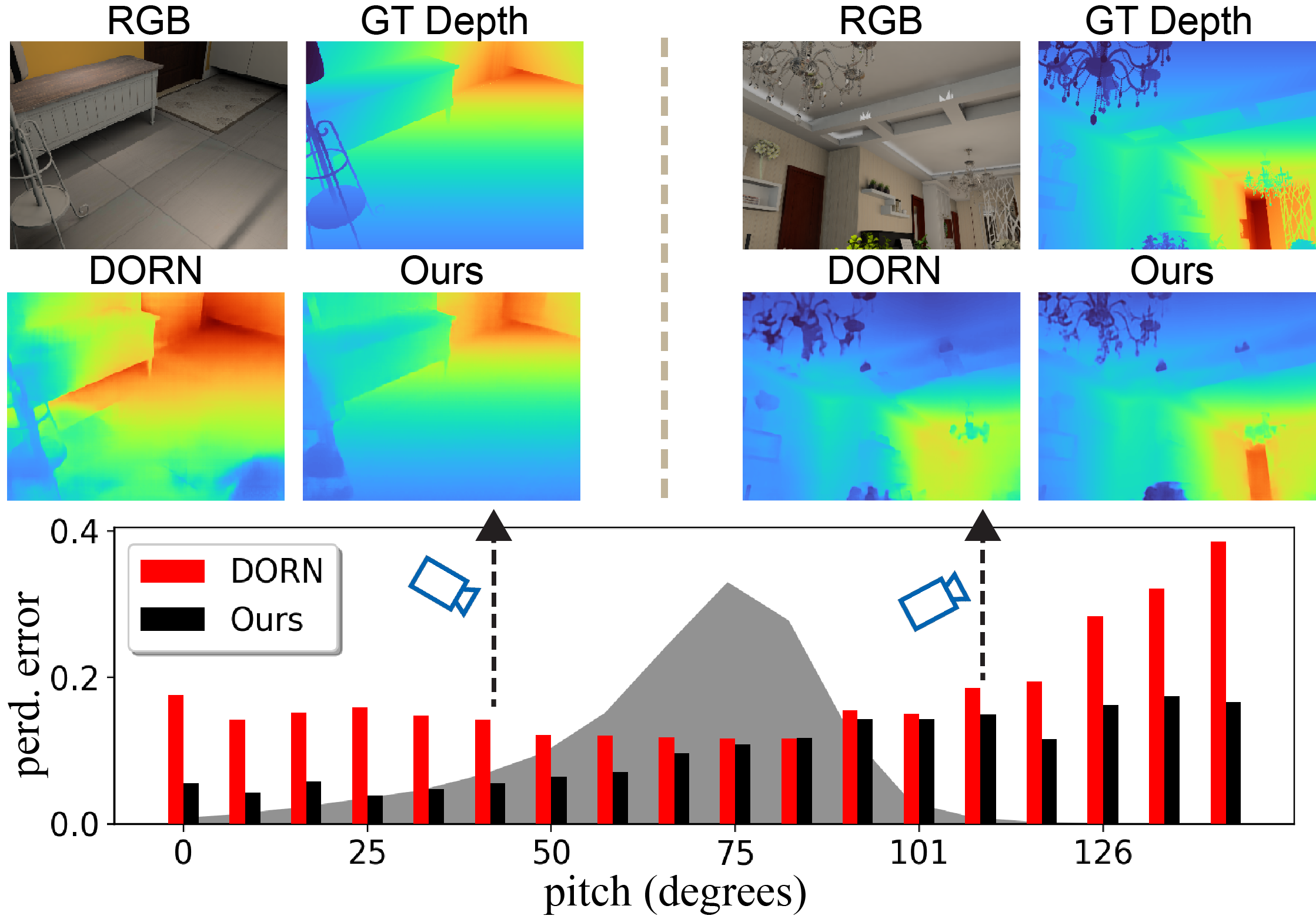} 
\vspace{-7mm}
\caption{\small
Contemporary monocular depth predictors, e.g., DORN~\cite{fu2018deep}, rely on large-scale training data which is naturally {\em biased}  w.r.t the distribution of camera poses (e.g., pitch angle distribution shown in {\color{shadegray}gray}). As a result, DORN makes unreliable predictions on test images captured with uncommon poses ({\color{red}red bars}), e.g., pitch angles $>$120$^{\circ}$.
To address this issue, we propose two novel techniques that drastically reduce prediction errors (cf. black bars) by leveraging 
perspective-aware data augmentation during training and known camera pose at test time.
Qualitative examples with more extreme camera pitch angles (top) show that incorporating our techniques leads to notable improvements.
}
\vspace{-5mm}
\label{fig:splashy-figure}
\end{figure}

However, an important overlooked fact is that the distribution of camera poses in training sets are naturally {\em biased}.
As a result, a learned depth predictor is unable to make reliable predictions on images captured from uncommon camera poses, as shown in Fig.~\ref{fig:splashy-figure}. 
Importantly, camera poses of testing examples may follow a different distribution from that in the training set. This will exacerbate prediction errors on images that are captured by cameras with uncommon poses relative to the training set.

\begin{figure}[t]
\centering
\includegraphics[width=0.99\linewidth]{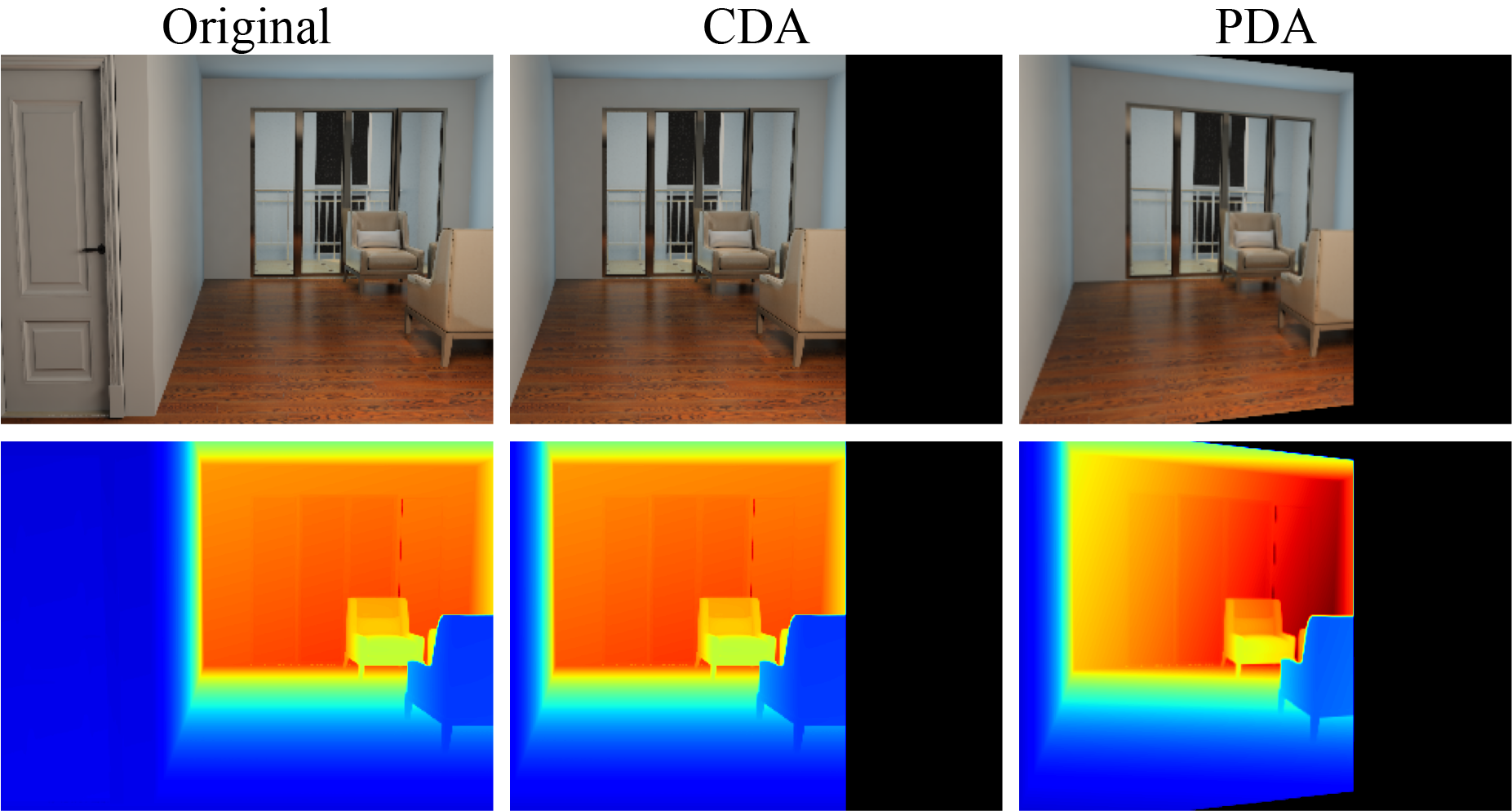}
\vspace{-2mm}
\caption{\small
{\bf Visual comparison of PDA and CDA}.
From the original example (left), conventional data augmentation (CDA) synthesizes a new example (middle) by randomly cropping a sub-region. 
It ignores camera pose information and will simply copy the depth values w.r.t the corresponding pixels.
In contrast, perspective-aware augmentation (PDA) simulates a rotation of the camera and synthesizes a new training example with
geometrically consistent depth values corresponding to the new camera pose  (right).}
\vspace{-4mm}
\label{fig: CDA vs PDA}
\end{figure}

{\bf Contributions}.
To this end,
we propose two novel approaches that significantly improve depth prediction under diverse test-time camera poses.
First, we introduce a simple \emph{perspective-aware data augmentation} (PDA) that synthesizes new geometrically consistent
training examples with more diverse viewpoints by perturbing the camera pose of existing samples.
In contrast, common data augmentation (CDA) methods such as random-crop, though widely adopted in prior work~\cite{fu2018deep, kong2018recurrent, zhang2019pattern, hu2019revisiting, yin2019enforcing, chen2019structure}, produce training examples where the resulting image
and target depth are inconsistent with the perspective geometry (Fig.~\ref{fig: CDA vs PDA}).
Second, we propose training a conditional depth predictor which utilizes the camera pose (e.g., acquired from IMU or other pose predictors) 
as a prior (CPP) when estimating depth. We propose an effective approach to encode CPP as an additional channel alongside the RGB input.
We find incorporating pose using CPP yields more accurate depth predictors that generalize much better under diverse test-time camera poses.

Through extensive experiments, we show that these techniques significantly improve depth prediction on images captured from uncommon and even never-before-seen camera poses.
Both techniques are general and broadly applicable to {\em any} network architecture. We show that incorporating them in recent state-of-the-art architectures improves their performance further. 
Lastly, we show that explicitly handling the biased camera pose distribution can improve the performance of a synthetically trained depth predictor when tested on real images. This highlights the importance that camera pose distribution plays in domain adaptation for 3D geometric prediction tasks.

\section{Related Work}

{\bf Monocular Depth Prediction}
and scene layout estimation have been greatly advanced since the seminal works~\cite{hoiem2005automatic, saxena20083}.
State-of-the-art approaches train increasingly sophisticated CNN-based predictors~\cite{liu2015learning, eigen2014depth, laina2016deeper}, utilize better training losses~\cite{fu2018deep, shin20193d, yin2019enforcing} and train on larger-scale training datasets~\cite{li2018megadepth, lasinger2019towards, zhang2017physically, zhao2020domain}. 

Surprisingly little attention has been paid to camera pose bias and out-of-distribution generalization.  The recent study of \cite{dijk2019neural} concluded that the learned depth predictors have a strong implicit bias causing them to perform poorly on test images when captured from differing camera poses. Our work systematically analyzes this pose bias/robustness in detail and offers technical contributions that improve generalization on test images captured under diverse camera poses.

{\bf Camera Pose Estimation} plays an essential role in many traditional 3D geometry vision problems such as SLAM~\cite{bailey2006simultaneous, sturm2012benchmark, kendall2015posenet} and 3D reconstruction~\cite{snavely2006photo, agarwal2011building}.
Predicting the relative camera pose between a pair of frames has been widely exploited to perform self-supervised learning of monocular depth prediction ~\cite{godard2017unsupervised, zhou2017unsupervised, ummenhofer2017demon}. Absolute camera pose is often represented implicitly in predictions 
of room layout~\cite{lee2017roomnet, zou2018layoutnet, xian2019uprightnet, workman2016horizon}. Closer to our work is~\cite{baradad2020height} 
which estimates the absolute camera pose (height,pitch,roll) in order to regularize depth predictions in world coordinates.

We explore the benefit of providing the camera pose as an additional {\em input} to depth prediction. In practical applications, such pose information 
may come from other sensors (e.g., pitch from IMU) or prior knowledge (e.g., camera mounted at a known height and pitch on an autonomous vehicle).
The work of \cite{he2018learning, cam-convs} encode camera intrinsics (e.g., focal length) as a part of the input for depth prediction, with a goal to learn a universal depth predictor that generalizes to images captured by different cameras. Similarly, we propose to encode camera extrinsic parameters (e.g., camera height) which we exploit for training better depth predictors that perform well on testing images captured with diverse camera extrinsic parameters.

{\bf Distribution Bias}.
Challenges of class imbalance and bias have been a widely discussed topic in the literature on classification and
recognition~\cite{he2009learning,zhu2014capturing,liu2019large}.
Distribution bias naturally exists in a training set, implying that some examples are underrepresented or few in number w.r.t some attributes (e.g., class labels). As a result, a trained model is unlikely to perform well on the underrepresented testing inputs.
Even worse, testing examples may come from out-of-distribution data~\cite{hendrycks2016baseline, liang2017enhancing, lee2017training}, 
meaning that the training set does not have similar examples. For example, in monocular depth prediction, training examples might be collected 
with cameras held vertical with minimal pitch and roll variations, but the testing scenario (e.g., AR/VR headset) might have potentially large 
variations in pitch and roll.


\begin{figure*}[t]
\centering
\includegraphics[width=0.99\textwidth]{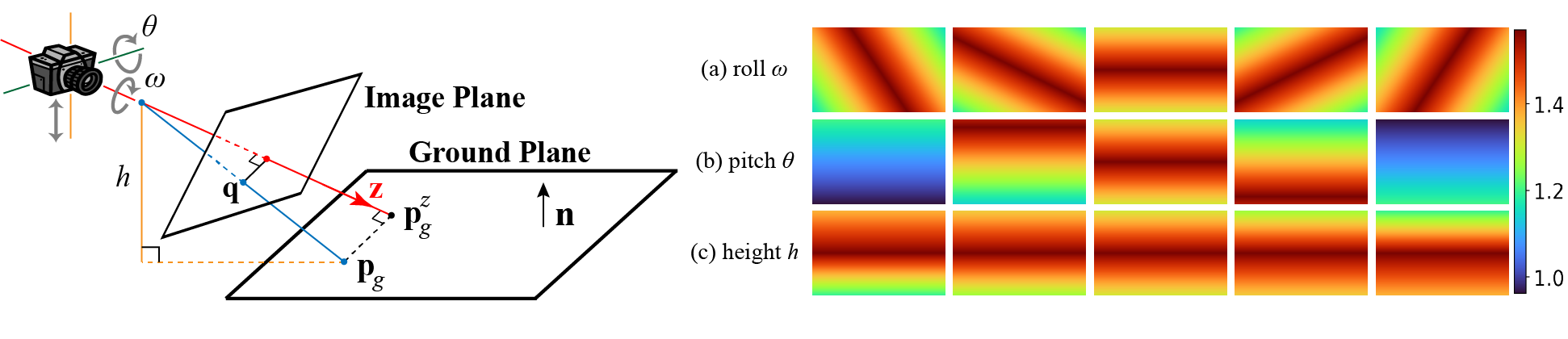}
\vspace{-7mm}
\caption{\small
{\bf Left:} We illustrate the proposed CPP which encodes camera pose
($\omega$, $\theta$, $h$) as a 2D image.
Intuitively, for a spatial coordinate $\q$ on the image plane (i.e., the encoding map), 
we find its physical point $\p_g$ on the ground plane, along the ray cast from the camera. Then we compute the pseudo depth value 
as the length from the camera to $\p_g^z$ which is the projection of $\p_g$ onto the depth direction $\z$ (red line) using Eqn.~\ref{eqn: final-project-cpp}.
This results in an encoded CPP map for a given camera pose.
{\bf Right:} We visualize some encoded  CPP maps by varying the $\omega$, $\theta$ and $h$ independently.
}
\vspace{-3mm}
\label{fig: camera pose encoding}
\end{figure*}

\section{Perspective-aware Data Augmentation}
\label{sec:PDA}

Due to the biased distribution of camera poses in the training set, some examples are underrepresented w.r.t camera poses.
To resolve this issue, we would like to augment the training data to cover these underrepresented poses. 

{\bf Resampling (RS)} the training data with replacement to enlarge the prevalence of uncommon poses in the train-set is perhaps the simplest approach.
However, this does not increase the diversity of the train-set because it cannot generate new training examples. Perhaps even worse, in practice,
training on repeated examples from uncommon camera poses
forces the model to weight them more (cf. overfitting), while sacrificing performance on other examples captured under common camera poses.

{\bf Conventional data augmentation (CDA)}, specifically random-cropping of the training examples, is widely used to enrich the training data in prior work~\cite{eigen2014depth, ladicky2014pulling, fu2018deep, kong2018recurrent, yin2019enforcing}.

While CDA seems to increase the diversity of the training set, its generated data could adversely affect depth prediction. Cropping a subregion from an original example naively copies depth values without considering the view-dependent nature of depth maps that depend on both the camera pose and scene geometry. Such a cropped image is equivalent to capturing another image of the same scene using a camera with an off-center principle point (Fig.~\ref{fig: CDA vs PDA}) and can make it difficult to train depth predictors~\cite{cam-convs}. 
In our work, we find CDA helps only if we crop sufficiently large regions, which presumably reduces this effect.

{\bf Perspective-aware data augmentation (PDA)} is our solution that augments training examples consisting of RGB images and depth maps. 
Given a training example, PDA first perturbs the camera pose, re-projects all pixels to a new image, and recomputes the depth values for the new image using the new camera pose and the original depth map.
Despite its simplicity, to the best of our knowledge, {\em no prior work exploits this idea for data augmentation during the training of a monocular depth predictor}.

Given the training image  ${\bf I}$, depth $\D$ and camera intrinsic matrix $\K$, we would like to synthesize a new image 
${\bf I}_{s}$ and depth ${\bf D}_{s}$ corresponding to a perturbed viewing direction.  Let ${\bf T}_{rel}$ be the relative
rotation between the two poses. Then for any point $\q = [u \ v]$ on ${\bf I}$, we compute the corresponding point $\q^{\prime}$ 
on the new image ${\bf I}_{s}$ via the homography:
\begin{equation}
    \q^{\prime}_h \sim \K {\bf T}_{rel} z \K^{-1} \q_h,
\end{equation}
where $z$ is the corresponding depth value of the point $\q$ from the original depth map $\D$; $\q_h$ and $\q^{\prime}_h$ are the homogeneous coordinates of points $\q$ and  $\q^{\prime}$, respectively.

Similarly, we compute the depth value for each pixel in the new depth map ${\bf D}_s(q^{\prime})$:
\begin{equation}
    z^{\prime} = {\bf v}^{\text{T}}_{proj} {\bf T}_{rel} z \K^{-1} \q_h,
\end{equation}
where ${\bf v}_{proj} = [0,0,1]^{\text{T}}$ is a unit vector pointed along the z-axis of the camera.

While the above demonstrates the computation of per-pixel depth values, in practice, we can compute the whole depth 
map $\D_s$ efficiently by using backward-warping and standard bilinear interpolation. For larger rotations, we
note that the synthesized views will have void regions on the boundary, as shown Fig.~\ref{fig: CDA vs PDA}.
This does not pose a problem for depth since we simply exclude those regions from the loss during training.
However, we find it helpful to pad the RGB void regions using values from the RGB images (using the ``reflection'' 
mode during warping).

We also considered applying PDA with camera translation.  However, this requires forward-warping and introduces
``holes'' in the synthesized result~\cite{masnou1998level} at disoccluion boundaries.
Handling disocclusions is still an open problem and out of our scope~\cite{buyssens2016depth, park2017transformation, luo2019disocclusion},
therefore, we choose to only augment camera rotations to avoid disocclusion artifacts and allow for efficient computation.

\section{Depth Prediction with Camera Pose Prior}
\label{sec:CPP}

Depth maps are {\em view-dependent representations} that depend on both the scene geometry and camera pose. 
The camera pose inherently provides \emph{prior knowledge} about the expected scene depth map. 
For example, knowing a camera is pointing down to the ground at one-meter height, we should expect a depth map of one-meter height roughly everywhere. 
Therefore, we are motivated to train a conditional depth predictor on camera pose as prior (CPP).

{\bf Camera pose} is a  six-degree-of-freedom (6DoF) vector that describes translation and rotation in 3D world coordinates.
In typical terrestrial man-made scenes, we consider a global coordinate with one axis pointing upwards (as specified by gravitational acceleration) and fix the origin along that axis to be 0 at the ground plane.
Since there is no unique origin along the two remaining axes, we assume that our camera pose prior should be uniform over translations parallel to the
ground plane.  
Similarly, there is no unique way to specify the orientation of the
axes parallel to the ground plane so our prior should necessarily be uniform
over rotations of the camera about the up-axis.
This leaves three degrees-of-freedom (3DoF): the height of the camera above the ground plane $h$, the pitch (angle relative to the up-axis) of the camera $\theta$, and any roll $\omega$ of the camera around its optical axis (Fig.~\ref{fig: camera pose encoding}).


\textbf{A naive encoding approach}.
We now consider using this 3DoF camera pose as a part of the input (along with RGB) to depth predictors.
To incorporate this as input into a CNN, inspired by the literature~\cite{cam-convs, you2019pseudo}, 
we convert 3DoF camera poses into 2D maps, which are concatenated with the RGB image as a whole input to learn the depth predictor. 
Naively, we can create three more channels of resolution $H\times W$, which copy values of roll $\omega$, pitch $\theta$ and height $h$, i.e.,
$\M_{\omega}[:,:]=\omega$, $\M_{\theta}[:,:]=\theta$ and $\M_{h}[:,:]=h$, respectively.  
However, the effect of pose on the depth distribution depends
strongly on the position in the image relative to the camera center so
translation-equivariant convolutions cannot fully exploit this encoding (except
by relying on boundary artifacts). This is supported by an experimental 
comparison showing this naive encoding is inferior
to our proposed \emph{CPP} encoding method, elaborated in the following.

\textbf{CPP encoding}
encodes the pose locally by assuming that the camera is
placed in an empty indoor scene with an infinite floor and ceiling.
Intuitively, it encodes the camera pose by intersecting rays from the camera center with predefined ground/ceiling planes and recording 
the depth, see Fig.~\ref{fig: camera pose encoding}.

Let $\p = [x \ y \ z]$ be a 3D point in the global coordinate, $\q = [u \ v]$ be a 2D point on the image plane whose homogeneous form is $\q_h$, $\n \in \mathbb{R}^3$ denotes the normal vector of ground planes, $C$ be the distance between two planes in the up direction.
 
The projection from 3D coordinates to 2D image plane is:
\begin{equation}
    \label{eqn: 2d-3d-projection}
    \lambda \q_h = \K \R^{-1} (\p - {\bf t}),
\end{equation}
where $\lambda$ is a scale factor; $\K$ is the intrinsic matrix; $\R \in SO(3)$ and ${\bf t} \in \mathbb{R}^3$ are rotation and translation matrices known from the camera pose, respectively. 
We compute the 3D point $\p$  where the ray shooting from the camera center through point $\q$ eventually intersects with planes or the horizon. However, $\lambda$ is an unknown scalar and we need extra constraints to compute it.
Taking ground plane as an example, we know the collections of 3D points intersecting with ground plane is $\{\p: \n^{\text{T}}\p = 0\}$. With this new constraint, we rearrange Eqn.~\ref{eqn: 2d-3d-projection} and multiply $\n^{\text{T}}$ on both sides of the equation:
\begin{align}
    \begin{split}
        \n^{\text{T}} \p &= \lambda \n^{\text{T}} \R \K^{-1} \q_h + \n^{\text{T}} {\bf t} \ \Rightarrow \
        \lambda = \frac{- \n^{\text{T}} {\bf t}}{\n^{\text{T}}\R\K^{-1}\q_h}
    \end{split}
\end{align}
Now, we plug the computed $\lambda$ back to Eqn.~\ref{eqn: 2d-3d-projection}, then the 3D point $\p_g$ that intersects the ground plane for $\q$ is:
\begin{equation}
    \p_g = \frac{- \n^{\text{T}} {\bf t}}{\n^{\text{T}}\R\K^{-1}\q_h} \R \K^{-1}\q_h + {\bf t}
\end{equation}
Similarly, with the constraint $\{\p: \n^{\text{T}}\p = C\}$, the 3D point $\p_c$ that intersects with the ceiling  plane for $\q$ is:
\begin{equation}
    {\p_c} = \frac{C - \n^{\text{T}} {\bf t}}{\n^{\text{T}}\R\K^{-1}\q_h} \R \K^{-1}\q_h + {\bf t}
\end{equation}
Once we have the 3D intersection point $\p$ for each point $\q$ on the image plane, we compute the projection on the camera $\z$ direction (i.e., the depth direction):
\begin{equation}
    \label{eqn: final-project-cpp}
    z(\p) = {\bf v}_{proj}^{\text{T}} \R^{-1}(\p - {\bf t}),
\end{equation}
where ${\bf v}_{proj}$ is the projection vector that computes the projection of a 3D point along the camera depth direction.
Finally, for each point $[u, v]$ in the encoding map $\M \in \mathbb{R}^{H \times W}$, we compute both $\p_g$ and $\p_c$ and take the maximum (positive) value as the pseudo depth value (Fig.~\ref{fig: camera pose encoding}):

\begin{equation}
\small
\M[u, v] = \max\{z(\p_g), z(\p_c)\}
\end{equation}
The encoding map $\M$ can have infinite values, e.g., when the ray shooting from the camera is parallel to the ground plane.  
To map values into a finite range, we apply the inverse tangent operator
$\tan^{-1}(\cdot)$ to obtain our final encoding $\M_{CPP} = \tan^{-1}(\M)$
which takes on values in the range $[\tan^{-1}(\min\{h, C-h\}), \frac{\pi}{2}]$. We visualize some CPP maps in Fig.~\ref{fig: camera pose encoding}-right.

To train a conditional depth prediction, we simply concatenate the CPP encoded map with the corresponding RGB as a four-channel input. 
This implies that our CPP encoding approach applies to {\em any} network architectures with a simple modification on the first convolution layer.

\section{Experiments}
\label{ssec:experiments}

We validate our methods through extensive experiments. Specifically, we aim to answer the following questions:\footnote{Answers: yes, no, yes, yes, yes.}

\begin{itemize}[noitemsep, topsep=-2pt, leftmargin=*]
\itemsep3pt
    \item Do our methods improve depth prediction on a test-set that has a different camera pose distribution from the train-set?
    \item Does resampling improve depth prediction on a test-set that has a different camera pose distribution?
    \item Do our methods improve depth estimation on testing images captured with  out-of-distribution camera poses?
    \item Does applying our methods to other state-of-the-art depth predictors improve their performance?
    \item Do our methods improve the performance of a depth predictor when training and testing on different datasets?
    \vspace{1mm}
\end{itemize}

\textbf{Datasets.}
In our work, 
we use two publicly available large-scale indoor-scene datasets, InteriorNet~\cite{interiornet} and ScanNet~\cite{ScanNet}
which come with ground-truth depth and camera pose.  Compared to other datasets such as~\cite{KITTI, Cordts2015Cvprw,song2017semantic},
these were selected because they illustrate a much wider variety of camera poses. InteriorNet~\cite{interiornet} consists 
of photo-realistic synthetic video sequences with randomly generated camera trajectories with a wide variety of pitch and 
height but minimal roll. ScanNet~\cite{ScanNet}  contains real-world RGBD videos of millions of views from 1,513 indoor 
scenes collected with substantial variation in pitch, roll, and height. For each dataset, we create train/test sets by randomly 
splitting \emph{scenes} into two disjoint sets with 85\%/15\% examples, respectively.
We use a stratified sampling approach to avoid picking adjacent video sequences in the same set.

\begin{figure}[t]
\centering
\includegraphics[width=0.95\linewidth]{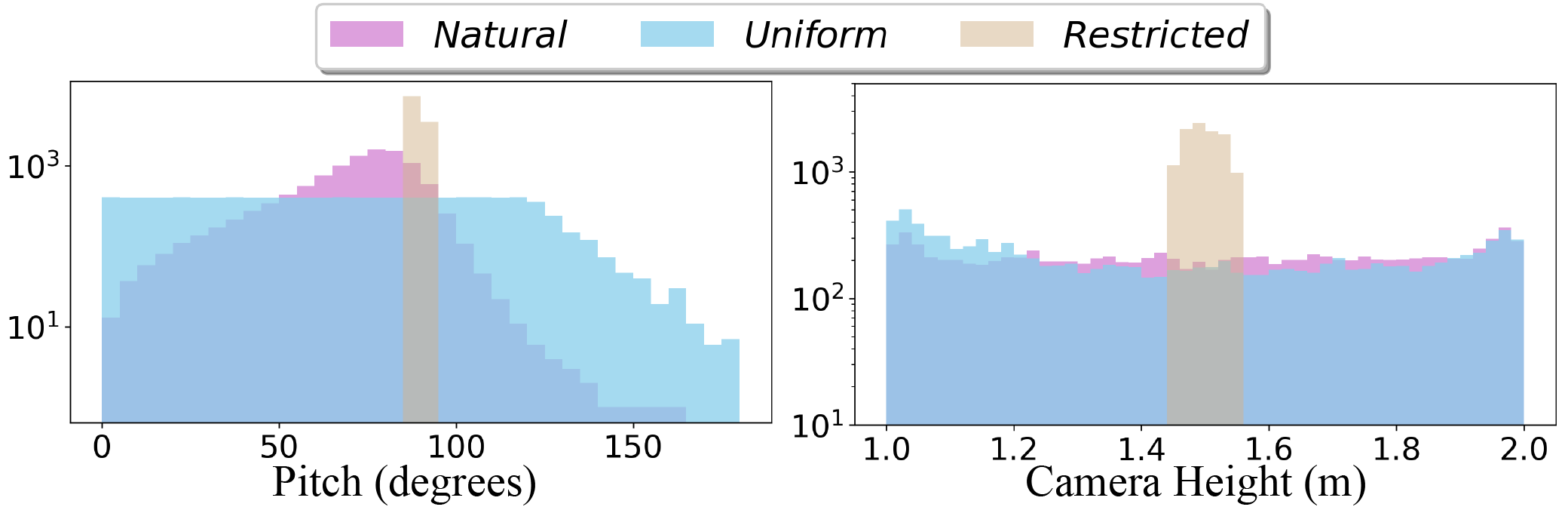}
\vspace{-2mm}
\caption{\small
Distribution of camera pitch and heights for three subsets of images from InteriorNet.  
From the \emph{Natural} subsect, we observe the dataset of InteriorNet does have a naturally biased distribution (esp. pitch). Please refer to the text on how we construct the three subsets.
}
\vspace{-4mm}
\label{fig: InteriorNet-statistics-figure}
\end{figure}

\textbf{Sampling.} 
To explore how the biased camera pose distribution affects depth learning and prediction, we sample three subsets from the
test/train set (Fig.~\ref{fig: InteriorNet-statistics-figure}) which each have 10k/1k images but with different distributions
of camera poses. 
\begin{itemize}[noitemsep, topsep=-2pt, leftmargin=*]
\item 
    \emph{Natural} selects samples at random to reflect the natural distribution of the dataset.
\item 
    \emph{Uniform} selects samples that simulate a uniform distribution over poses with priority: pitch$\rightarrow$roll$\rightarrow$height. Concretely, we quantize the range of camera pitch/roll/ height into equal-size bins and sample an approximately equal number of examples in each bin. 
\item 
    \emph{Restricted} samples images within a narrow range: camera pitch $\theta \in [85^\circ, 95^\circ]$ and height $h \in [1.45, 1.55]$ (meters). While InteriorNet does not have roll variations, ScanNet does: roll $\omega \in [-5^\circ, 5^\circ]$. 
    We create \emph{Restricted} to particularly study how depth predictor performs on testing images captured with out-of-distribution camera poses.
\vspace{1mm}
\end{itemize}

\begin{table}[t]
\centering
\caption{\small
{\bf Within \& cross-distribution evaluation}. 
In each dataset, we train depth predictors on their \emph{Natural} train-sets and evaluate  on both \emph{Natural} and \emph{Uniform} test-sets. 
We apply different methods to a Vanilla model.
All models use the same network architecture.
Vanilla performs poorly in cross-distribution evaluation (cf. {\em Natural}-test vs. {\em Uniform}-test), demonstrating that the biased camera pose distribution affects the training of depth predictors. 
As expected, RS hurts depth prediction compared to Vanilla.
In contrast, CPP and PDA show better performance; jointly applying them performs the best (i.e., ``Both''). 
Finally, comparing alternative methods to our CPP (vs. Native)  and PDA (vs. RS and CDA) shows the merits of our methods.
}
\vspace{-1mm}
\begin{adjustbox}{max width=\linewidth}
{
\centering
\begin{tabular}{l| c c | c c }
\hline
\multirow{3}{*}{Models} &  \multicolumn{2}{c|}{\textit{Natural-Test-Set}} & \multicolumn{2}{c}{\textit{Uniform-Test-Set}} \\
& \multicolumn{1}{c|}{\cellcolor{col1} \texttt{$\downarrow$ better}} &  \multicolumn{1}{c|}{\cellcolor[rgb]{0.0,0.8,1.0} \texttt{$\uparrow$ better}}  &
\multicolumn{1}{c|}{\cellcolor{col1} \texttt{$\downarrow$ better}} &  \multicolumn{1}{c}{\cellcolor[rgb]{0.0,0.8,1.0} \texttt{$\uparrow$ better}} \\
& \multicolumn{1}{c|}{\cellcolor{col1}  \small Abs$^r$/Sq$^r$/RMS-log} & \multicolumn{1}{c|}{\cellcolor[rgb]{0.0,0.8,1.0} $\delta^1$ \ / \ $\delta^2$}
& \multicolumn{1}{c|}{\cellcolor{col1} \small Abs$^r$/Sq$^r$/RMS-log} & \multicolumn{1}{c}{\cellcolor[rgb]{0.0,0.8,1.0} $\delta^1$ \ / \ $\delta^2$} \\
\hline
\multicolumn{5}{c}{\cellcolor{lightgrey}\tt InteriorNet} \\
Vanilla
& \multicolumn{1}{c}{.154 / .148 / .229} & \multicolumn{1}{c|}{.803 / .945}
& \multicolumn{1}{c}{.183 / .146 / .250} & \multicolumn{1}{c}{.724 / .926} \\
\hline
+ RS
& \multicolumn{1}{c}{.192 / .203 / .267} & \multicolumn{1}{c|}{.726 / .918 }
& \multicolumn{1}{c}{.210 / .174 / .272} & \multicolumn{1}{c}{.661 / .906} \\
+ CDA
& \multicolumn{1}{c}{.142 / .137 / .222} & \multicolumn{1}{c|}{.825 / .950}
& \multicolumn{1}{c}{.172 / .125 / .238} & \multicolumn{1}{c}{.738 / .934} \\
+ PDA
& \multicolumn{1}{c}{.138 / .123 / .207} & \multicolumn{1}{c|}{.834 / .957}
& \multicolumn{1}{c}{.168 / .122 / .229} & \multicolumn{1}{c}{.757 / .942} \\
\hline
+ Naive
& \multicolumn{1}{c}{.132 / .145 / .219} & \multicolumn{1}{c|}{.835 / .944}
& \multicolumn{1}{c}{.137 / .116 / .213} & \multicolumn{1}{c}{.810 / .944} \\
+ CPP
& \multicolumn{1}{c}{.108 / .120 / .199} & \multicolumn{1}{c|}{.872 / .958}
& \multicolumn{1}{c}{.106 / .088 / .183} & \multicolumn{1}{c}{.876 / .961} \\
\hline
+ Both
& \multicolumn{1}{c}{\bf .095 / .101 / .180} & \multicolumn{1}{c|}{\bf .898 / .966}
& \multicolumn{1}{c}{\bf .091 / .069 / .161} & \multicolumn{1}{c}{\bf .903 / .973} \\
\hline
\multicolumn{5}{c}{\cellcolor{lightgrey}\tt ScanNet} \\
Vanilla
& \multicolumn{1}{c}{.125 / .068 / .186} & \multicolumn{1}{c|}{.837 / .962}
& \multicolumn{1}{c}{.177 / .121 / .265} & \multicolumn{1}{c}{.711 / .928} \\
\hline
+ RS
& \multicolumn{1}{c}{.216 / .158 / .279} & \multicolumn{1}{c|}{.619 / .889}
& \multicolumn{1}{c}{.218 / .168 / .300} & \multicolumn{1}{c}{.630 / .881} \\
+ CDA
& \multicolumn{1}{c}{.116 / .062 / .179} & \multicolumn{1}{c|}{.853 / .964}
& \multicolumn{1}{c}{.174 / .121 / .264} & \multicolumn{1}{c}{.727 / .922} \\
+ PDA
& \multicolumn{1}{c}{.115 / .059 / .171} & \multicolumn{1}{c|}{.860 / .970}
& \multicolumn{1}{c}{.166 / .110 / .248} & \multicolumn{1}{c}{.752 / .938} \\
\hline
+ Naive
& \multicolumn{1}{c}{.120 / .069 / .184} & \multicolumn{1}{c|}{.846 / .959}
& \multicolumn{1}{c}{.173 / .127 / .255} & \multicolumn{1}{c}{.755 / .923} \\
+ CPP
& \multicolumn{1}{c}{.108 / .060 / .171} & \multicolumn{1}{c|}{.871 / .965}
& \multicolumn{1}{c}{.154 / .106 / .239} & \multicolumn{1}{c}{.781 / .943} \\
\hline
+ Both
& \multicolumn{1}{c}{\bf .102 / .052 / .160} & \multicolumn{1}{c|}{\bf .882 / .973}
& \multicolumn{1}{c}{\bf .143 / .097 / .230} & \multicolumn{1}{c}{\bf .809 / .952} \\
\hline
\end{tabular}
}
\end{adjustbox}
\label{tab:cross distributions}
\end{table}

\begin{figure}[t]
\centering
\includegraphics[width=\linewidth]{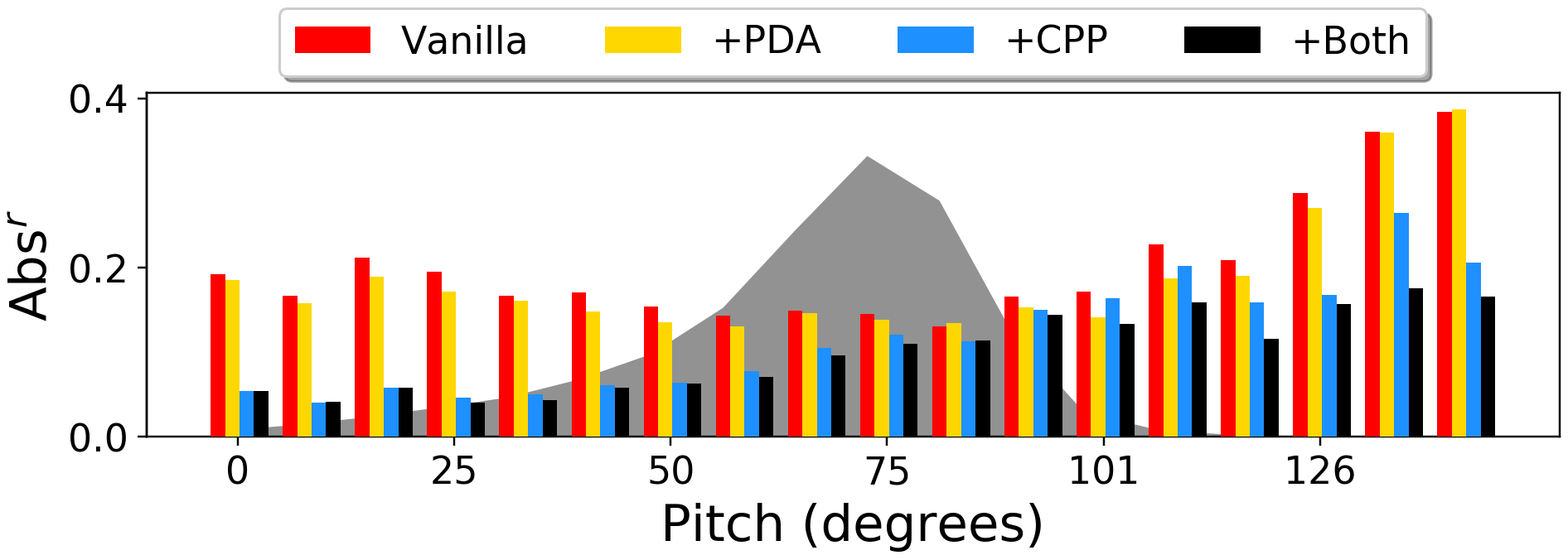}
\vspace{-6mm}
\caption{\small
Breakdown analysis of depth prediction w.r.t pitch. 
The background shade denotes the camera pose distribution w.r.t pitch.
Clearly, both PDA and CPP improve depth prediction in underrepresented camera poses.
Surprisingly, CPP remarkably boosts depth prediction, while applying both CPP and PDA achieves the best performance ``everywhere''.
}
\vspace{-3mm}
\label{fig: cross-distribution-figure}
\end{figure}

\textbf{Evaluation Metrics.} 
There are several evaluation metrics widely used in the literature~\cite{eigen2014depth, fu2018deep, yin2019enforcing},
including absolute relative difference (Abs$^r$), squared relative difference (Sq$^r$), root mean squared log error (RMS-log), and accuracy with a relative error threshold of $\delta^k < 1.25^k$, $i = 1, 2$.

\begin{figure*}[t]
\centering
\includegraphics[width=0.99\textwidth]{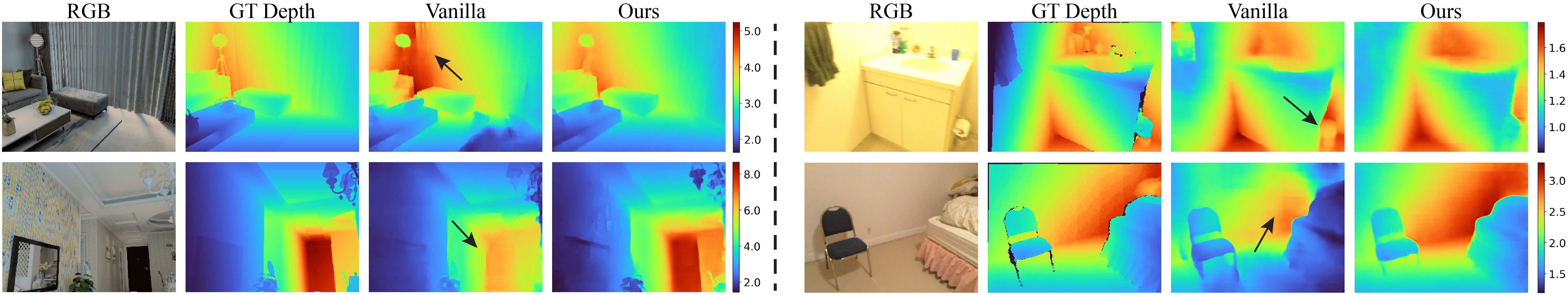}
\vspace{-3mm}
\caption{\small
Qualitative comparison between Vanilla and our method (using both CPP and PDA) on random testing images from InteriorNet (left) and ScanNet (right). Depth maps for each example are shown with the same colorbar range.
 Notably, the images are captured under some uncommon camera angles relative to the training pose distribution 
(Fig.~\ref{fig: cross-distribution-figure}).
The Vanilla model seems to make erroneous predictions w.r.t the overall scale of the depth. In contrast, by applying CPP and PDA, the new model (``ours'') produces visually improved results.}
\vspace{-2mm}
\label{fig:vis-demo}
\end{figure*}

\textbf{Implementation}.
We use a standard UNet structured model to perform most of our experimental analysis~\cite{zhao2020domain, zhao2019geometry, zheng2018t2net}. 
We also demonstrate that our techniques apply to other depth predictor architectures (Section~\ref{ssec:sota-comparison}).
Unless specified, all models are trained with the L1 loss and applied random left-right flip augmentation during training.

While we initially learn the conditional depth predictor using the {\em true} camera pose, we also tested encoding a 
{\em predicted} camera pose in Section~\ref{ssec:generalization}.
For predicting camera pose, we train a pose predictor with the ResNet18 architecture~\cite{he2016deep} to directly regresses 
to camera pitch, roll, and height. 
We resize all images and depth maps to $240 \times 320$, and adjust camera intrinsic parameters (for CPP encoding) accordingly.  
We use PyTorch~\cite{paszke2017automatic} to train all the models for 200 epochs on a single Titan Xp GPU.
We use the Adam optimizer~\cite{adam} with a learning rate 1e-3, 
and coefficients 0.5 and 0.999 for computing the running averages of gradient and its square.

For CPP encoding, we set the ceiling height $C=3$ meters. We have studied different settings of $C$, but find little difference (details in the supplement).
For PDA, we randomly perturb camera pose within $[-0.1, 0.1]$ radius angle jointly w.r.t pitch, roll, and yaw; we also ablate the perturbation scale in Section~\ref{ssec:ablation}.

\subsection{Within \& Cross-Distribution Evaluation}
\label{ssec:within-cross-dist}

We start with initial experiments designed to reveal how bias in camera pose statistics affects depth predictor performance
(Fig.~\ref{fig: InteriorNet-statistics-figure}).
The experiments also explore the design choices of our methods and validate their effectiveness.
Specifically, we train depth predictors on the \emph{Natural} train-sets, and test them on both \emph{Natural} and \emph{Uniform} test-sets.
We apply a sequence of different modifications to a ``Vanilla'' baseline model, i.e., a UNet-based predictor. 
Table~\ref{tab:cross distributions} lists detailed comparisons, and Fig.~\ref{fig:vis-demo} shows qualitative comparisons.

The Vanilla model degrades notably when evaluated on a test-set that has a different pose distribution, i.e., from {\em Natural} to {\em Uniform}, showing the clear influence from the biased distribution of camera poses.
In terms of data augmentation, simply resampling the training data (RS) hurts performance (cf.  Vanilla and ``+RS'').
Moreover, our PDA outperforms CDA, demonstrating the importance of synthesizing geometric-aware training examples using the corresponding camera and the scene geometry in depth prediction.
As for camera pose encoding, our CPP encoding method clearly outperforms the Naive method,
Importantly, jointly applying CPP and PDA  performs the best on both within-distribution ({\em Natural})  and cross-distribution ({\em Uniform}) test-sets.
Finally, we breakdown the performance in Fig.~\ref{fig: cross-distribution-figure} to analyze when (i.e., w.r.t pitch angle)  PDA and CPP improve depth prediction. Generally, both of them help reduce prediction errors on testing images captured with underrepresented camera poses, while CPP yields the largest benefits. Applying both achieves the best performance ``{\em everywhere}''.

\subsection{Out-of-Distribution Evaluation}
\label{ssec:ood-eval}

We now study how our methods help when training depth predictors on a train-set which have a rather restricted range of camera poses.
Specifically, we train models on the {\em Restricted} train-sets of InteriorNet and ScanNet, respectively, and test the models on their {\em Natural} test-sets.
This setup is synthetic and unlikely to be a real-world scenario, but it allows for exclusive analysis of depth predictors when tested on images captured with out-of-distribution or never-before-seen camera poses.

Table~\ref{tab:extrapolations} lists detailed comparisons.
Vanilla model performs poorly when tested on {\em Natural} test-set.
CPP slightly improves prediction, but PDA helps a lot.
CDA (i.e., random-crop as augmentation) also helps, presumably owing to more diverse training examples, but underperforms our PDA. 
This further confirms the importance of  generating geometric-aware new training examples for the given scene and camera in depth prediction.
Under expectation, applying both PDA and CPP performs the best.

\begin{table}[t]
\centering
\caption{\small
{\bf Out-of-distribution evaluation}. We train depth predictors on the \emph{Restricted} train-sets and test on both \emph{Restricted} and \emph{Natural} test-sets of each datasets. 
Vanilla model performs poorly on \emph{Natural} test-sets, clearly showing the challenge of depth prediction on images captured under novel/never-before-seen camera poses.
CPP slightly improves performance, but PDA helps more.
CDA also improves performance presumably because it synthesizes more training examples, but underperforms PDA.
As expected, jointly applying both CPP and PDA achieves the best performance on both {\em Restricted} and {\em Natural} test-sets.}
\vspace{-3mm}
\begin{adjustbox}{max width=\linewidth}
{
\centering
\begin{tabular}{l| c c | c c }
\hline
\multirow{3}{*}{Models} &  \multicolumn{2}{c|}{\textit{Restricted-Test-Set}} & \multicolumn{2}{c}{\textit{Natural-Test-Set}} \\
& \multicolumn{1}{c|}{\cellcolor{col1} \texttt{$\downarrow$ better}} &  \multicolumn{1}{c|}{\cellcolor[rgb]{0.0,0.8,1.0} \texttt{$\uparrow$ better}}  &
\multicolumn{1}{c|}{\cellcolor{col1} \texttt{$\downarrow$ better}} &  \multicolumn{1}{c}{\cellcolor[rgb]{0.0,0.8,1.0} \texttt{$\uparrow$ better}} \\
& \multicolumn{1}{c|}{\cellcolor{col1}  \small Abs$^r$/Sq$^r$/RMS-log} & \multicolumn{1}{c|}{\cellcolor[rgb]{0.0,0.8,1.0} $\delta^1$ \ / \ $\delta^2$}
& \multicolumn{1}{c|}{\cellcolor{col1} \small Abs$^r$/Sq$^r$/RMS-log} & \multicolumn{1}{c}{\cellcolor[rgb]{0.0,0.8,1.0} $\delta^1$ \ / \ $\delta^2$} \\
\hline
\multicolumn{5}{c}{\cellcolor{lightgrey}\tt InteriorNet} \\
Vanilla
& \multicolumn{1}{c}{.150 / .234 / .296} & \multicolumn{1}{c|}{.819 / .916}
& \multicolumn{1}{c}{.265 / .368 / .412} & \multicolumn{1}{c}{.538 / .767} \\
+ CPP
& \multicolumn{1}{c}{.149 / .237 / .301} & \multicolumn{1}{c|}{.820 / .915}
& \multicolumn{1}{c}{.264 / .350 / .397} & \multicolumn{1}{c}{.548 / .782} \\
+ CDA
& \multicolumn{1}{c}{.139 / .228 / .286} & \multicolumn{1}{c|}{.830 / .922}
& \multicolumn{1}{c}{.227 / .279 / .336} & \multicolumn{1}{c}{.637 / .845} \\
+ PDA
& \multicolumn{1}{c}{.136 / .178 / .239} & \multicolumn{1}{c|}{.833 / .935}
& \multicolumn{1}{c}{.237 / .266 / .295} & \multicolumn{1}{c}{.652 / .878} \\
+ Both
& \multicolumn{1}{c}{\bf .131 / .170 / .237} & \multicolumn{1}{c|}{\bf .839 / .936}
& \multicolumn{1}{c}{\bf .216 / .231 / .286} & \multicolumn{1}{c}{\bf .666 / .894} \\
\hline
\multicolumn{5}{c}{\cellcolor{lightgrey}\tt ScanNet} \\
Vanilla
& \multicolumn{1}{c}{.177 / .131 / .259} & \multicolumn{1}{c|}{.705 / .897}
& \multicolumn{1}{c}{.317 / .304 / .390} & \multicolumn{1}{c}{.444 / .748} \\
+ CPP
& \multicolumn{1}{c}{.169 / .125 / .258} & \multicolumn{1}{c|}{.726 / .897}
& \multicolumn{1}{c}{.301 / .283 / .384} & \multicolumn{1}{c}{.464 / .756} \\
+ CDA
& \multicolumn{1}{c}{.163 / .121 / .259} & \multicolumn{1}{c|}{.724 / .904}
& \multicolumn{1}{c}{.289 / .266 / .371} & \multicolumn{1}{c}{.467 / .771} \\
+ PDA
& \multicolumn{1}{c}{.160 / .112 / .244} & \multicolumn{1}{c|}{.729 / .914}
& \multicolumn{1}{c}{.283 / .251 / .353} & \multicolumn{1}{c}{.493 / .795} \\
+ Both
& \multicolumn{1}{c}{\bf .155 / .108 / .228} & \multicolumn{1}{c|}{\bf .731 / .918}
& \multicolumn{1}{c}{\bf .277 / .245 / .348} & \multicolumn{1}{c}{\bf .504 / .804} \\
\hline
\end{tabular}
}
\end{adjustbox}
\vspace{-2mm}
\label{tab:extrapolations}
\end{table}

\subsection{Applicability to Other Predictor Networks}
\label{ssec:sota-comparison}

CPP and PDA are general and applicable to different model architectures.
We study applying them to training two well-established depth predictors: DORN~\cite{fu2018deep} and VNL~\cite{yin2019enforcing}.
Compared to the Vanilla model, both predictors adopt different network architectures~\cite{xie2017aggregated, he2016deep} and different loss functions.
They convert continuous depth values to discrete bins and model depth prediction as a classification problem.
Moreover, VNL incorporates a virtual surface normal loss which provides a geometric-aware regularization during training.
We implement DORN and VNL using publicly-available third-party code. 
We train and test all models on the {\em Natural} train/test-sets, in InteriorNet and ScanNet, respectively.

Table~\ref{tab:sota comparisons} lists detailed results, which are comparable to the \emph{Natural-Test-Set} column in Table~\ref{tab:cross distributions}.
Consistent with previous experiments, both CPP and PDA improve the performance further based on DORN and VNL.
As our CPP and PDA improve other depth predictors as a general approach, we suggest using them in future research of monocular depth estimation.

\begin{table}[t]
\centering

\caption{\small
{\bf Applicability to other predictor architectures}. 
In each dataset, we train state-of-the-art
depth predictors (DORN~\cite{fu2018deep} and VNL~\cite{yin2019enforcing}) by optionally applying our CPP and PDA approaches.
All models are trained/tested on the \emph{Natural} train/test-sets per dataset.
Clearly, both CPP and PDA boost their performance.
}
\vspace{-3mm}
\begin{adjustbox}{max width=\linewidth}
{
\centering
\begin{tabular}{l| c c | c c }
\hline
\multirow{3}{*}{Models} &  \multicolumn{2}{c|}{\textit{\tt InteriorNet}} & \multicolumn{2}{c}{\textit{\tt ScanNet}} \\
& \multicolumn{1}{c|}{\cellcolor{col1} \texttt{$\downarrow$ better}} &  \multicolumn{1}{c|}{\cellcolor[rgb]{0.0,0.8,1.0} \texttt{$\uparrow$ better}}  &
\multicolumn{1}{c|}{\cellcolor{col1} \texttt{$\downarrow$ better}} &  \multicolumn{1}{c}{\cellcolor[rgb]{0.0,0.8,1.0} \texttt{$\uparrow$ better}} \\
& \multicolumn{1}{c|}{\cellcolor{col1}  \small Abs$^r$/Sq$^r$/RMS-log} & \multicolumn{1}{c|}{\cellcolor[rgb]{0.0,0.8,1.0} $\delta^1$ \ / \ $\delta^2$}
& \multicolumn{1}{c|}{\cellcolor{col1} \small Abs$^r$/Sq$^r$/RMS-log} & \multicolumn{1}{c}{\cellcolor[rgb]{0.0,0.8,1.0} $\delta^1$ \ / \ $\delta^2$} \\
\hline
DORN
& \multicolumn{1}{c}{.128 / .126 / .218} & \multicolumn{1}{c|}{.854 / .957}
& \multicolumn{1}{c}{.112 / .065 / .197} & \multicolumn{1}{c}{.856 / .959} \\
+ CPP
& \multicolumn{1}{c}{.098 / .105 / .197} & \multicolumn{1}{c|}{.899 / .962}
& \multicolumn{1}{c}{.108 / .062 / .191} & \multicolumn{1}{c}{.875 / .960} \\
+ PDA
& \multicolumn{1}{c}{.115 / .112 / .207} & \multicolumn{1}{c|}{.867 / .958}
& \multicolumn{1}{c}{.110 / .068 / .195} & \multicolumn{1}{c}{.857 / .963} \\
+ Both
& \multicolumn{1}{c}{\bf .085 / .088 / .124} & \multicolumn{1}{c|}{\bf .912 / .971}
& \multicolumn{1}{c}{\bf .093 / .049 / .153} & \multicolumn{1}{c}{\bf .903 / .979} \\
\hline
VNL
& \multicolumn{1}{c}{.131 / .140 / .235} & \multicolumn{1}{c|}{.853 / .954}
& \multicolumn{1}{c}{.115 / .069 / .205} & \multicolumn{1}{c}{.855 / .960} \\
+ CPP
& \multicolumn{1}{c}{.101 / .120 / .207} & \multicolumn{1}{c|}{.893 / .960}
& \multicolumn{1}{c}{.112 / .064 / .195} & \multicolumn{1}{c}{.871 / .961} \\
+ PDA
& \multicolumn{1}{c}{.117 / .115 / .210} & \multicolumn{1}{c|}{.861 / .959}
& \multicolumn{1}{c}{.110 / .065 / .199} & \multicolumn{1}{c}{.855 / .962} \\
+ Both
& \multicolumn{1}{c}{\bf .086 / .085 / .131} & \multicolumn{1}{c|}{\bf .909 / .970}
& \multicolumn{1}{c}{\bf .095 / .051 / .157} & \multicolumn{1}{c}{\bf .901 / .980} \\
\hline
\end{tabular}
}
\end{adjustbox}
\vspace{-3mm}
\label{tab:sota comparisons}
\end{table}

\begin{figure*}[t]
\centering
\includegraphics[width=\textwidth]{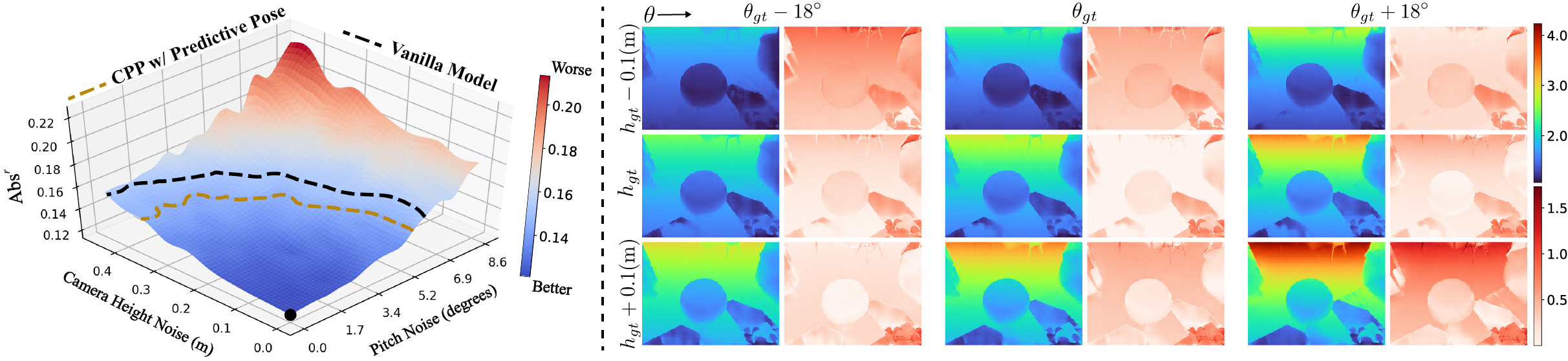}
\vspace{-6mm}
\caption{\small
{\bf Left:} We plot depth prediction error (Abs$^r$) w.r.t different levels of noise in camera height and pitch. 
We apply CPP to train/test a depth predictor (based on Vanilla) on InteriorNet \emph{Natural} train/test-set.
For a given noise level $\delta$, the trained model makes depth predictions using a CPP map computed with a perturbed camera pose, e.g., the pitch is sampled from
$\theta_{gt}$$+$$\U[-\delta, \delta]$.  
The black dot at the origin stands for the (best) performance using the true camera pose (i.e., no noises are presented in pitch and height).
The dashed lines represent the average performance levels for the Vanilla and CPP with predictive poses.
{\bf Right}: We visualize depth prediction by CPP model when encoding perturbed camera pitch
angles $\theta_{gt}$$\pm$$18^\circ$ and heights $h_{gt}$$\pm$$0.1$(m).
CPP model predicts shallower depth when both pitch and camera height decrease (i.e., camera is tilted down or translated closer to the floor).  This qualitatively confirms that the camera pose prior induces a meaningful shift in the estimator.  
The corresponding RGB image and ground-truth depth appear in Fig.~\ref{fig: blind prediction}.
}
\vspace{-2mm}
\label{fig:resilience-study}
\end{figure*}

\subsection{Synthetic-to-Real Generalization}
\label{ssec:generalization}

Previous experiments have validated that addressing the biased camera pose distribution helps train a  depth predictor that works better on another 
test-time distribution, or more generally another domain.
Here we evaluate performance in the presence of substantial synthetic-to-real domain
shift which includes both low-level shifts (e.g., local material appearance) and high-level shifts (novel objects, layouts and camera poses)~\cite{zhao2020domain}.
Specifically, we synthetically train a depth predictor (on InteriorNet \emph{Natural} train-set) and test it on real images (ScanNet \emph{Natural} and \emph{Uniform} test-sets).
We also consider a more practical scenario that one does not have access to the true camera pose but instead must rely on the predicted poses. To this end, we train a camera pose predictor on ScanNet  \emph{Natural} test-set to predict camera pitch, height and roll for a given RGB image. Then, we perform CPP encoding with the predictive pose, i.e., CPP$_{pred}$

Table~\ref{tab:cross dataset} lists detailed setups and results; we summarize the salient conclusions.
Vanilla achieves worse performance compared to Table~\ref{tab:cross distributions}, showing a clear domain gap between the two datasets.
Applying CDA hurts the performance, presumably because the generated training data by CDA
disobey the projective geometry relations between scene geometry and camera model and pose (cf. Section~\ref{sec:PDA}).
In contrast, our PDA helps, but CPP improves even more notably. Applying CPP with the predictive pose (CPP$_{pred}$) achieves a remarkable performance boost over PDA, suggesting that using predictive poses, or more generally exploiting camera poses during training, is quite valuable in depth prediction. We analyze in the next section how the model is resilient to the errors in predicted poses. Lastly, using the true camera pose in CPP performs the best.

\begin{table}[t]
\centering
\caption{\small
{\bf Mitigating distribution bias of camera poses improves synthetic-to-real domain adaptation}.
We train depth predictors synthetically on InteriorNet (\emph{Natural} train-set) and test them on real-world images  from ScanNet \textit{Natural} and \textit{Uniform} test-sets. 
This is a typical setup for synthetic-to-real domain adaptation in the context of depth prediction.
Interestingly, we find that CDA hurts the performance, presumably because the generated training examples by CDA do not obey the relations among camera model, scene geometry and camera pose, and hence do not necessarily help training a generalizable depth predictor.
In contrast, our PDA helps synthetic-to-real generalization and applying CPP improves further.
Importantly, applying CPP with predictive poses (CPP$_{pred}$) achieves a remarkable performance boost, whereas using the true camera pose in CPP performs the best.
}
\vspace{-2mm}
\begin{adjustbox}{max width=\linewidth}
{
\centering
\begin{tabular}{l | c c | c c}
\hline
\multirow{3}{*}{\shortstack[l]{Methods}} 
&  \multicolumn{2}{c|}{\textit{Natural-Test-Set}} & \multicolumn{2}{c}{\textit{Uniform-Test-Set}} \\
& \multicolumn{1}{c|}{\cellcolor{col1} \texttt{$\downarrow$ better}} &  \multicolumn{1}{c}{\cellcolor[rgb]{0.0,0.8,1.0} \texttt{$\uparrow$ better}}
& \multicolumn{1}{c|}{\cellcolor{col1} \texttt{$\downarrow$ better}} &  \multicolumn{1}{c}{\cellcolor[rgb]{0.0,0.8,1.0} \texttt{$\uparrow$ better}} \\
& \multicolumn{1}{c|}{\cellcolor{col1} \small Abs$^r$/Sq$^r$/RMS-log} & \multicolumn{1}{c}{\cellcolor[rgb]{0.0,0.8,1.0} $\delta^1$ \ / \ $\delta^2$}
& \multicolumn{1}{c|}{\cellcolor{col1} \small Abs$^r$/Sq$^r$/RMS-log} & \multicolumn{1}{c}{\cellcolor[rgb]{0.0,0.8,1.0} $\delta^1$ \ / \ $\delta^2$} \\
\hline
{Vanilla}
& \multicolumn{1}{c}{.242 / .204 / .315} & \multicolumn{1}{c|}{.570 / .852}
& \multicolumn{1}{c}{.305 / .259 / .364} & \multicolumn{1}{c}{.457 / .797} \\
{CDA} & \multicolumn{1}{c}{.246 / .205 / .321} & \multicolumn{1}{c|}{.568 / .843} & \multicolumn{1}{c}{.316 / .288 / .391} & \multicolumn{1}{c}{.433 / .771} \\
{PDA}
& \multicolumn{1}{c}{.239 / .198 / .311} & \multicolumn{1}{c|}{.575 / .857}
& \multicolumn{1}{c}{.298 / .252 / .359} &  \multicolumn{1}{c}{.461 / .804} \\
PDA+CPP$_{pred}$
& \multicolumn{1}{c}{.219 / .190 / .305} & \multicolumn{1}{c|}{.586 / .866}
& \multicolumn{1}{c}{.273 / .231 / .346} & \multicolumn{1}{c}{.548 / .835} \\
PDA+CPP
& \multicolumn{1}{c}{\bf .208 / .170 / .282} & \multicolumn{1}{c|}{\bf .677 / .893}
& \multicolumn{1}{c}{\bf .245 / .228 / .315} & \multicolumn{1}{c}{\bf .620 / .858} \\
\hline
\end{tabular}
}
\end{adjustbox}
\vspace{-4mm}
\label{tab:cross dataset}
\end{table}

\subsection{Further Discussion and Ablation Study}
\label{ssec:ablation}

\begin{figure}[t]
\centering
\includegraphics[width=0.99\linewidth]{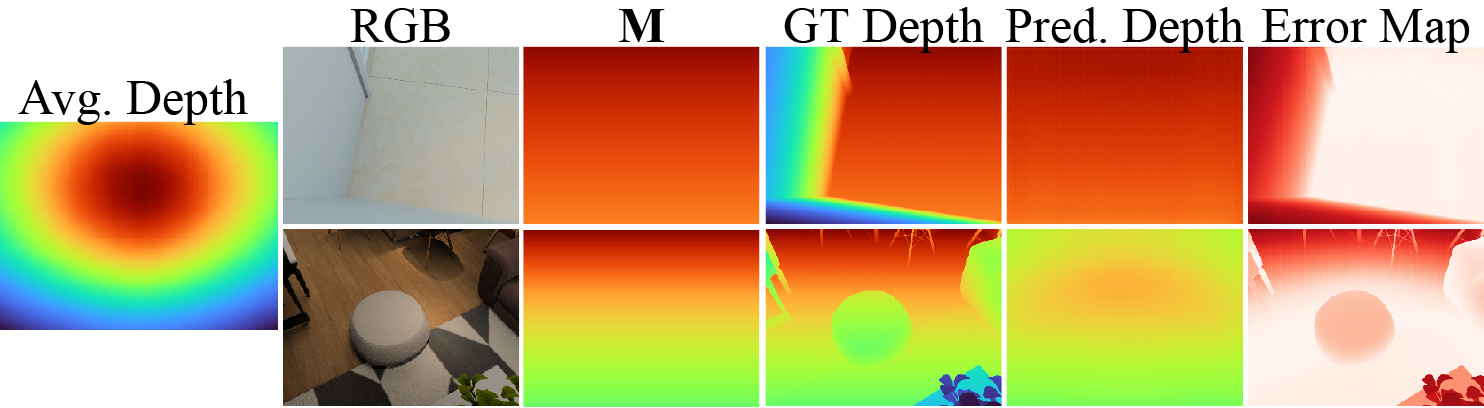}
\vspace{-3mm}
\caption{\small
Camera pose alone provides a strong depth prior even for ``blind'' depth prediction. Specifically, over the InteriorNet \emph{Natural} train-set, we train a depth predictor solely on the CPP encoded maps $\M$ {\em without} RGB as input. For visual comparison, we compute the average depth map (shown left).
We visualize depth predictions on two random examples.
All the depth maps are visualized with the same colormap range.
Perhaps not surprisingly, $\M$ presents nearly the true depth in floor areas, suggesting that camera pose alone does provide strong prior depth information for these scenes.
}
\vspace{-5mm}
\label{fig: blind prediction}
\end{figure}

\textbf{``Blind'' depth prediction without RGB}.
To characterize the prior knowledge carried by camera poses in terms of depth prediction, we train a ``blind predictor'' on the InteriorNet \emph{Natural} train-set, taking as input only the CPP encoded maps of camera poses \emph{without} RGB images. 
For comparison, we compute an average depth map over the whole \emph{Natural} train-set.  
We qualitatively compare results on two testing examples in Fig.~\ref{fig: blind prediction} (quantitative results in the supplement).
Visually, encoding camera pose alone using CPP reliably provides depth estimate on floor regions. This intuitively explains that using camera pose does serve as strong prior knowledge of scene depth.

\textbf{Resilience to noisy camera pose}.
Camera pose estimates (e.g., from IMU sensors) are potentially noisy, and the predicted camera poses are undoubtedly erroneous (cf. the previous experiment using predicted poses in CPP).
We study how resilient CPP is to test-time errors in the camera pose.
Specifically, when testing a trained model with CPP,
we randomly perturb the camera poses of all testing examples up to a pre-defined scale, and measure the overall performance as a function of prediction error w.r.t noise added to the true camera pitch $\theta_{gt}$ and height $h_{gt}$, as shown in Fig.~\ref{fig:resilience-study}.
We find that CPP outperforms the vanilla model even with significant misspecification of the pose (e.g., height error $<$ 0.3m, pitch error $<$ 5 degrees).

\begin{figure}[t]
\centering
\includegraphics[width=\linewidth]{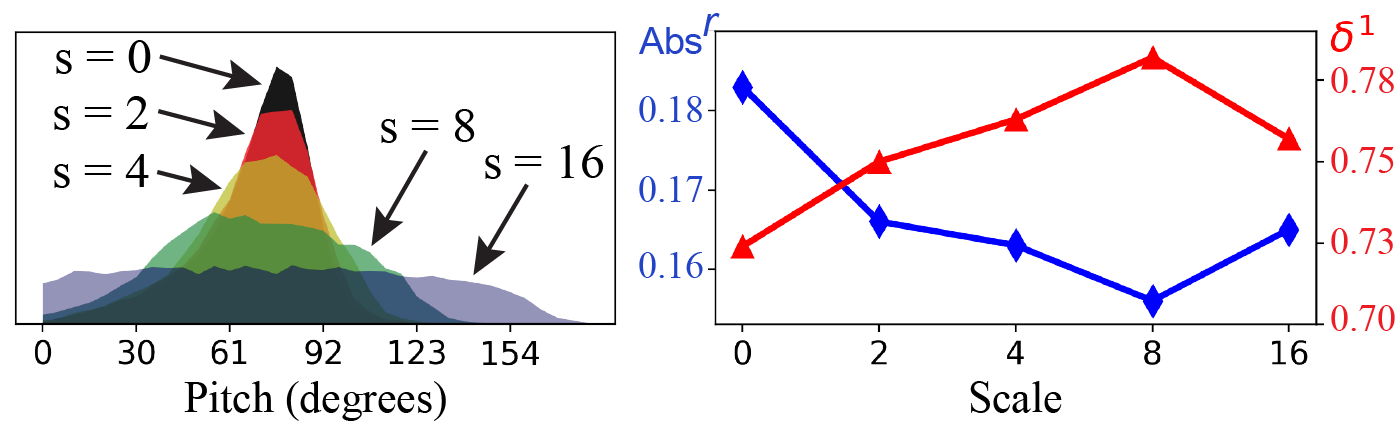}
\vspace{-2mm}
\caption{\small
During training on InteriorNet \emph{Natural} train-set, we randomly perturb camera pose to generate new training examples.
We specify the scale of the perturbation $s=\{0, 2, 4, 8, 16\}$, meaning that, when $s=2$, we randomly perturb pitch/roll/yaw angles by adding a perturbation within $[-s*5^{\circ}, s*5^{\circ}]$.
{\bf Left}: Applying more larger scale PDA ``flattens'' camera pose distribution of the whole training set.
{\bf Right}: We test each of the trained models on the InteriorNet \emph{Uniform} test-set. We find applying more intense PDA consistently improves depth prediction until $s=16$ (i.e., rotating at most 80$^\circ$), presumably when very large void regions are introduced in the synthesized training examples (Fig.~\ref{fig: CDA vs PDA}).}
\vspace{-5mm}
\label{fig: aug scale}
\end{figure}

\textbf{Augmentation scales in PDA}.
We ablate the augmentation scale in PDA during training depth predictors, as detailed in Fig.~\ref{fig: aug scale}.
Perhaps surprisingly, applying a larger scale PDA consistently improves depth prediction until a very large perturbation (i.e., rotating at most 80$^\circ$), presumably when very large void regions are introduced in the synthesized training examples (Fig.~\ref{fig: CDA vs PDA}).

\section{Conclusion}
While large-scale datasets allow for end-to-end training of monocular depth predictors, 
we find the training sets naturally biased w.r.t distribution of camera poses. As a result, trained predictors fail to make reliable depth predictions for testing examples captured under uncommon camera poses.
We mitigate this bias with two novel methods, perspective-aware data augmentation (PDA) and camera pose prior encoding (CPP). 
We show that applying both our methods improves depth prediction on images captured under uncommon or never-before-seen camera poses.
Moreover, our methods are general and readily applicable to other depth predictors, which can perform better when trained with PDA and CPP, suggesting using them in the future research of monocular depth estimation.

\noindent\textbf{Acknowledgements} We thank anonymous reviewers for their valuable suggestions. This research was supported by NSF grants IIS-1813785, IIS-1618806, a research gift from Qualcomm, and a hardware donation from NVIDIA.

{\small
\bibliographystyle{ieee_fullname}
\bibliography{egbib}

\begin{thebibliography}{10}\itemsep=-1pt

\bibitem{agarwal2011building}
Sameer Agarwal, Yasutaka Furukawa, Noah Snavely, Ian Simon, Brian Curless,
  Steven~M Seitz, and Richard Szeliski.
\newblock Building rome in a day.
\newblock {\em Communications of the ACM}, 54(10):105--112, 2011.

\bibitem{bailey2006simultaneous}
Tim Bailey and Hugh Durrant-Whyte.
\newblock Simultaneous localization and mapping (slam): Part ii.
\newblock {\em IEEE robotics \& automation magazine}, 2006.

\bibitem{baradad2020height}
Manel Baradad and Antonio Torralba.
\newblock Height and uprightness invariance for 3d prediction from a single
  view.
\newblock In {\em CVPR}, 2020.

\bibitem{bian2019unsupervised}
Jiawang Bian, Zhichao Li, Naiyan Wang, Huangying Zhan, Chunhua Shen, Ming-Ming
  Cheng, and Ian Reid.
\newblock Unsupervised scale-consistent depth and ego-motion learning from
  monocular video.
\newblock In {\em Advances in Neural Information Processing Systems}, 2019.

\bibitem{buyssens2016depth}
Pierre Buyssens, Olivier Le~Meur, Maxime Daisy, David Tschumperl{\'e}, and
  Olivier L{\'e}zoray.
\newblock Depth-guided disocclusion inpainting of synthesized rgb-d images.
\newblock {\em IEEE Transactions on Image Processing}, 2016.

\bibitem{chen2019structure}
Xiaotian Chen, Xuejin Chen, and Zheng-Jun Zha.
\newblock Structure-aware residual pyramid network for monocular depth
  estimation.
\newblock {\em arXiv preprint arXiv:1907.06023}, 2019.

\bibitem{Cordts2015Cvprw}
Marius Cordts, Mohamed Omran, Sebastian Ramos, Timo Scharw{\"a}chter, Markus
  Enzweiler, Rodrigo Benenson, Uwe Franke, Stefan Roth, and Bernt Schiele.
\newblock The cityscapes dataset.
\newblock In {\em CVPR Workshop on The Future of Datasets in Vision}, 2015.

\bibitem{ScanNet}
Angela Dai, Angel~X Chang, Manolis Savva, Maciej Halber, Thomas Funkhouser, and
  Matthias Nie{\ss}ner.
\newblock Scannet: Richly-annotated 3d reconstructions of indoor scenes.
\newblock In {\em CVPR}, 2017.

\bibitem{dijk2019neural}
Tom~van Dijk and Guido~de Croon.
\newblock How do neural networks see depth in single images?
\newblock In {\em ICCV}, 2019.

\bibitem{eigen2015predicting}
David Eigen and Rob Fergus.
\newblock Predicting depth, surface normals and semantic labels with a common
  multi-scale convolutional architecture.
\newblock In {\em ICCV}, 2015.

\bibitem{eigen2014depth}
David Eigen, Christian Puhrsch, and Rob Fergus.
\newblock Depth map prediction from a single image using a multi-scale deep
  network.
\newblock In {\em Advances in Neural Information Processing Systems}, 2014.

\bibitem{cam-convs}
Jose~M Facil, Benjamin Ummenhofer, Huizhong Zhou, Luis Montesano, Thomas Brox,
  and Javier Civera.
\newblock Cam-convs: camera-aware multi-scale convolutions for single-view
  depth.
\newblock In {\em CVPR}, 2019.

\bibitem{fu2018deep}
Huan Fu, Mingming Gong, Chaohui Wang, Kayhan Batmanghelich, and Dacheng Tao.
\newblock Deep ordinal regression network for monocular depth estimation.
\newblock In {\em CVPR}, 2018.

\bibitem{KITTI}
Andreas Geiger, Philip Lenz, and Raquel Urtasun.
\newblock Are we ready for autonomous driving? the kitti vision benchmark
  suite.
\newblock In {\em CVPR}, 2012.

\bibitem{godard2017unsupervised}
Cl{\'e}ment Godard, Oisin Mac~Aodha, and Gabriel~J Brostow.
\newblock Unsupervised monocular depth estimation with left-right consistency.
\newblock In {\em CVPR}, 2017.

\bibitem{guo2018learning}
Xiaoyang Guo, Hongsheng Li, Shuai Yi, Jimmy Ren, and Xiaogang Wang.
\newblock Learning monocular depth by distilling cross-domain stereo networks.
\newblock In {\em ECCV}, 2018.

\bibitem{he2009learning}
Haibo He and Edwardo~A Garcia.
\newblock Learning from imbalanced data.
\newblock {\em IEEE Transactions on knowledge and data engineering}, 2009.

\bibitem{he2016deep}
Kaiming He, Xiangyu Zhang, Shaoqing Ren, and Jian Sun.
\newblock Deep residual learning for image recognition.
\newblock In {\em CVPR}, pages 770--778, 2016.

\bibitem{he2018learning}
Lei He, Guanghui Wang, and Zhanyi Hu.
\newblock Learning depth from single images with deep neural network embedding
  focal length.
\newblock {\em IEEE Transactions on Image Processing}, 2018.

\bibitem{hendrycks2016baseline}
Dan Hendrycks and Kevin Gimpel.
\newblock A baseline for detecting misclassified and out-of-distribution
  examples in neural networks.
\newblock {\em arXiv preprint arXiv:1610.02136}, 2016.

\bibitem{hoiem2005automatic}
Derek Hoiem, Alexei~A Efros, and Martial Hebert.
\newblock Automatic photo pop-up.
\newblock In {\em ACM SIGGRAPH 2005 Papers}, pages 577--584. 2005.

\bibitem{hu2019revisiting}
Junjie Hu, Mete Ozay, Yan Zhang, and Takayuki Okatani.
\newblock Revisiting single image depth estimation: Toward higher resolution
  maps with accurate object boundaries.
\newblock In {\em 2019 IEEE Winter Conference on Applications of Computer
  Vision (WACV)}, 2019.

\bibitem{kendall2015posenet}
Alex Kendall, Matthew Grimes, and Roberto Cipolla.
\newblock Posenet: A convolutional network for real-time 6-dof camera
  relocalization.
\newblock In {\em ICCV}, 2015.

\bibitem{adam}
Diederik~P Kingma and Jimmy Ba.
\newblock Adam: A method for stochastic optimization.
\newblock {\em arXiv preprint arXiv:1412.6980}, 2014.

\bibitem{kong2018recurrent}
Shu Kong and Charless~C Fowlkes.
\newblock Recurrent scene parsing with perspective understanding in the loop.
\newblock In {\em CVPR}, 2018.

\bibitem{ladicky2014pulling}
Lubor Ladicky, Jianbo Shi, and Marc Pollefeys.
\newblock Pulling things out of perspective.
\newblock In {\em CVPR}, 2014.

\bibitem{laina2016deeper}
Iro Laina, Christian Rupprecht, Vasileios Belagiannis, Federico Tombari, and
  Nassir Navab.
\newblock Deeper depth prediction with fully convolutional residual networks.
\newblock In {\em 2016 Fourth international conference on 3D vision (3DV)},
  2016.

\bibitem{lasinger2019towards}
Katrin Lasinger, Ren{\'e} Ranftl, Konrad Schindler, and Vladlen Koltun.
\newblock Towards robust monocular depth estimation: Mixing datasets for
  zero-shot cross-dataset transfer.
\newblock {\em arXiv preprint arXiv:1907.01341}, 2019.

\bibitem{lee2017roomnet}
Chen-Yu Lee, Vijay Badrinarayanan, Tomasz Malisiewicz, and Andrew Rabinovich.
\newblock Roomnet: End-to-end room layout estimation.
\newblock In {\em ICCV}, 2017.

\bibitem{lee2017training}
Kimin Lee, Honglak Lee, Kibok Lee, and Jinwoo Shin.
\newblock Training confidence-calibrated classifiers for detecting
  out-of-distribution samples.
\newblock {\em arXiv preprint arXiv:1711.09325}, 2017.

\bibitem{interiornet}
Wenbin Li, Sajad Saeedi, John McCormac, Ronald Clark, Dimos Tzoumanikas, Qing
  Ye, Yuzhong Huang, Rui Tang, and Stefan Leutenegger.
\newblock Interiornet: Mega-scale multi-sensor photo-realistic indoor scenes
  dataset.
\newblock {\em arXiv preprint arXiv:1809.00716}, 2018.

\bibitem{li2018megadepth}
Zhengqi Li and Noah Snavely.
\newblock Megadepth: Learning single-view depth prediction from internet
  photos.
\newblock In {\em CVPR}, 2018.

\bibitem{liang2017enhancing}
Shiyu Liang, Yixuan Li, and Rayadurgam Srikant.
\newblock Enhancing the reliability of out-of-distribution image detection in
  neural networks.
\newblock {\em arXiv preprint arXiv:1706.02690}, 2017.

\bibitem{liu2015learning}
Fayao Liu, Chunhua Shen, Guosheng Lin, and Ian Reid.
\newblock Learning depth from single monocular images using deep convolutional
  neural fields.
\newblock {\em IEEE Transactions on Pattern Analysis and Machine Intelligence},
  2015.

\bibitem{liu2019large}
Ziwei Liu, Zhongqi Miao, Xiaohang Zhan, Jiayun Wang, Boqing Gong, and Stella~X
  Yu.
\newblock Large-scale long-tailed recognition in an open world.
\newblock In {\em CVPR}, 2019.

\bibitem{luo2019disocclusion}
Guibo Luo, Yuesheng Zhu, Zhenyu Weng, and Zhaotian Li.
\newblock A disocclusion inpainting framework for depth-based view synthesis.
\newblock {\em IEEE Transactions on Pattern Analysis and Machine Intelligence},
  2019.

\bibitem{masnou1998level}
Simon Masnou and J-M Morel.
\newblock Level lines based disocclusion.
\newblock In {\em Proceedings 1998 International Conference on Image
  Processing}, 1998.

\bibitem{park2017transformation}
Eunbyung Park, Jimei Yang, Ersin Yumer, Duygu Ceylan, and Alexander~C Berg.
\newblock Transformation-grounded image generation network for novel 3d view
  synthesis.
\newblock In {\em CVPR}, 2017.

\bibitem{paszke2017automatic}
Adam Paszke, Sam Gross, Soumith Chintala, Gregory Chanan, Edward Yang, Zachary
  DeVito, Zeming Lin, Alban Desmaison, Luca Antiga, and Adam Lerer.
\newblock Automatic differentiation in pytorch.
\newblock 2017.

\bibitem{saxena20083}
Ashutosh Saxena, Sung~H Chung, and Andrew~Y Ng.
\newblock 3-d depth reconstruction from a single still image.
\newblock {\em International Journal of Computer Vision}, 76(1):53--69, 2008.

\bibitem{shin20193d}
Daeyun Shin, Zhile Ren, Erik~B Sudderth, and Charless~C Fowlkes.
\newblock 3d scene reconstruction with multi-layer depth and epipolar
  transformers.
\newblock In {\em ICCV}, 2019.

\bibitem{snavely2006photo}
Noah Snavely, Steven~M Seitz, and Richard Szeliski.
\newblock Photo tourism: exploring photo collections in 3d.
\newblock In {\em ACM Siggraph 2006 Papers}, pages 835--846. 2006.

\bibitem{song2017semantic}
Shuran Song, Fisher Yu, Andy Zeng, Angel~X Chang, Manolis Savva, and Thomas
  Funkhouser.
\newblock Semantic scene completion from a single depth image.
\newblock In {\em CVPR}, 2017.

\bibitem{sturm2012benchmark}
J{\"u}rgen Sturm, Nikolas Engelhard, Felix Endres, Wolfram Burgard, and Daniel
  Cremers.
\newblock A benchmark for the evaluation of rgb-d slam systems.
\newblock In {\em 2012 IEEE/RSJ International Conference on Intelligent Robots
  and Systems}, pages 573--580. IEEE, 2012.

\bibitem{ummenhofer2017demon}
Benjamin Ummenhofer, Huizhong Zhou, Jonas Uhrig, Nikolaus Mayer, Eddy Ilg,
  Alexey Dosovitskiy, and Thomas Brox.
\newblock Demon: Depth and motion network for learning monocular stereo.
\newblock In {\em CVPR}, 2017.

\bibitem{workman2016horizon}
Scott Workman, Menghua Zhai, and Nathan Jacobs.
\newblock Horizon lines in the wild.
\newblock {\em arXiv preprint arXiv:1604.02129}, 2016.

\bibitem{xian2019uprightnet}
Wenqi Xian, Zhengqi Li, Matthew Fisher, Jonathan Eisenmann, Eli Shechtman, and
  Noah Snavely.
\newblock Uprightnet: geometry-aware camera orientation estimation from single
  images.
\newblock In {\em ICCV}, 2019.

\bibitem{xie2017aggregated}
Saining Xie, Ross Girshick, Piotr Doll{\'a}r, Zhuowen Tu, and Kaiming He.
\newblock Aggregated residual transformations for deep neural networks.
\newblock In {\em CVPR}, pages 1492--1500, 2017.

\bibitem{yin2019enforcing}
Wei Yin, Yifan Liu, Chunhua Shen, and Youliang Yan.
\newblock Enforcing geometric constraints of virtual normal for depth
  prediction.
\newblock In {\em ICCV}, 2019.

\bibitem{you2019pseudo}
Yurong You, Yan Wang, Wei-Lun Chao, Divyansh Garg, Geoff Pleiss, Bharath
  Hariharan, Mark Campbell, and Kilian~Q Weinberger.
\newblock Pseudo-lidar++: Accurate depth for 3d object detection in autonomous
  driving.
\newblock {\em arXiv preprint arXiv:1906.06310}, 2019.

\bibitem{zhang2017physically}
Yinda Zhang, Shuran Song, Ersin Yumer, Manolis Savva, Joon-Young Lee, Hailin
  Jin, and Thomas Funkhouser.
\newblock Physically-based rendering for indoor scene understanding using
  convolutional neural networks.
\newblock In {\em CVPR}, 2017.

\bibitem{zhang2019pattern}
Zhenyu Zhang, Zhen Cui, Chunyan Xu, Yan Yan, Nicu Sebe, and Jian Yang.
\newblock Pattern-affinitive propagation across depth, surface normal and
  semantic segmentation.
\newblock In {\em CVPR}, 2019.

\bibitem{zhao2019geometry}
Shanshan Zhao, Huan Fu, Mingming Gong, and Dacheng Tao.
\newblock Geometry-aware symmetric domain adaptation for monocular depth
  estimation.
\newblock In {\em CVPR}, 2019.

\bibitem{zhao2020domain}
Yunhan Zhao, Shu Kong, Daeyun Shin, and Charless Fowlkes.
\newblock Domain decluttering: Simplifying images to mitigate synthetic-real
  domain shift and improve depth estimation.
\newblock In {\em CVPR}, 2020.

\bibitem{zheng2018t2net}
Chuanxia Zheng, Tat-Jen Cham, and Jianfei Cai.
\newblock T2net: Synthetic-to-realistic translation for solving single-image
  depth estimation tasks.
\newblock In {\em ECCV}, 2018.

\bibitem{zhou2017unsupervised}
Tinghui Zhou, Matthew Brown, Noah Snavely, and David~G Lowe.
\newblock Unsupervised learning of depth and ego-motion from video.
\newblock In {\em CVPR}, 2017.

\bibitem{zhu2014capturing}
Xiangxin Zhu, Dragomir Anguelov, and Deva Ramanan.
\newblock Capturing long-tail distributions of object subcategories.
\newblock In {\em CVPR}, 2014.

\bibitem{zou2018layoutnet}
Chuhang Zou, Alex Colburn, Qi Shan, and Derek Hoiem.
\newblock Layoutnet: Reconstructing the 3d room layout from a single rgb image.
\newblock In {\em CVPR}, 2018.

\end{thebibliography}
}

\clearpage
\appendix
\addcontentsline{toc}{section}{Appendices}

\begin{center}
{\bf \Large Appendices}
\end{center}

\emph{
In the supplementary document, we provide additional experimental results to further support our findings, as well as details of our experiments and more visualizations. Below is the outline.
}
\begin{itemize} [noitemsep, topsep=-1pt, leftmargin=*]
\item 
    {\bf Section~\ref{ssec:cpp_encoding}:  Handling infinite values in CPP encoding.}
    In the main paper, we apply the inverse tangent operation to deal with infinite values in the CPP encoded maps. We study an alternative based on a simple clipping operation.
\item
    {\bf Section~\ref{ssec:distance_C_study}:  Hyperparameter analysis in CPP encoding.} 
    Our CPP encoding method has a hyperparameter $C$ which is the distance between the upper and lower planes (i.e., a ceiling and ground plane).
    We study how $C$ affects depth prediction performance.
\item
    {\bf Section~\ref{ssec:blind}:  Quantitative results of blind predictions.} We show the quantitative results of blind prediction to better understand how CPP helps capture prior knowledge for depth prediction.
\item
    {\bf Section~\ref{ssec:aug}:  Further study of PDA augmentation scales.} We provide a thorough study between the depth predictor performance and PDA augmentation scales.
\item 
    {\bf Section~\ref{ssec:fixed_pitch_height}: Further study of camera height and rotation in CPP encoding.} We compare the performance of CPP on InteriorNet (with nearly fixed roll) by either encoding the true pitch or camera height while fixing the other one.
\item 
    {\bf Section~\ref{ssec:predicted poses}: CPP Encoding with Predicted Poses.} Considering the scenario where test time camera poses are not always available, we show experimental results of CPP encoding with predicted poses during evaluations.
\item 
    {\bf Section~\ref{ssec:impl}: Additional details in experiments.} We present more experimental details such as RGB and depth preprocessing steps, training the camera pose prediction models, our evaluation protocol, and the ScanNet camera pose distribution.
\item
    {\bf More Visual Results.} We include more depth prediction visualizations of different methods in Fig.~\ref{fig:results-1} and \ref{fig:results-2}.
\end{itemize}

\section{Handling Infinite Values in CPP Encoding}
\label{ssec:cpp_encoding}
At the last step in computing CPP encoded maps $\M_{CPP}$, we apply the inverse tangent operator to eliminate the infinity values (happens when the ray shooting from camera is parallel to the ground plane) and maps the values of $\M$ (i.e. $\M_{CPP} = \tan^{-1}(\M)$) to the range $[\tan^{-1}(\min\{h, C-h\}), \frac{\pi}{2}]$.
However, the inverse tangent operator is not the only choice. We provide an ablation study that replaces the $\tan^{-1}(\cdot)$ with a clipping operation.

\begin{figure}[t]
    \centering
    \includegraphics[width=0.99\linewidth]{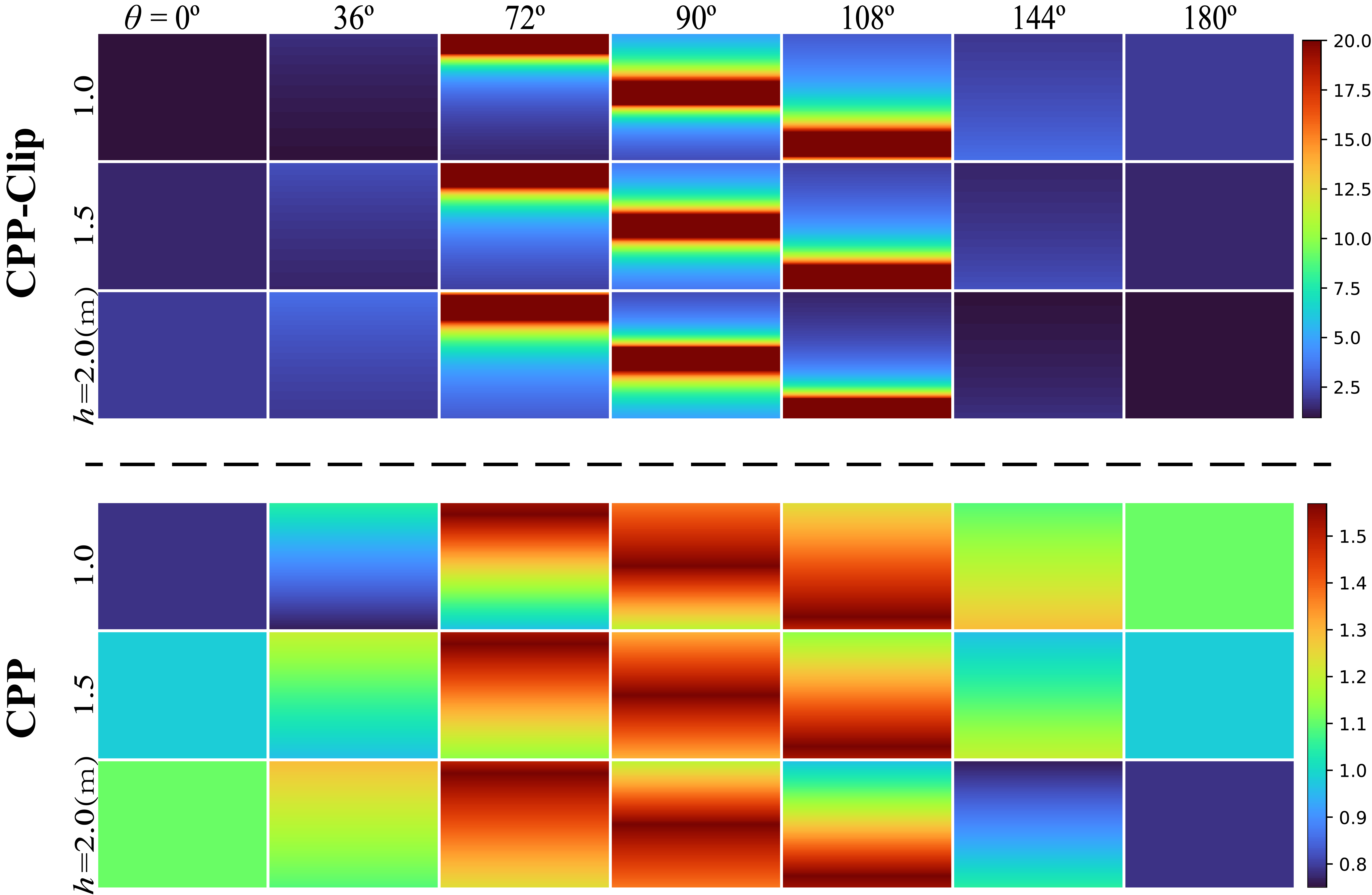}
    \vspace{-3mm}
    \caption{\small
    Visual comparisons of encoded maps of CPP and CPP-Clip with different pitch $\theta$ and camera height $h$. We set the threshold $\tau$=20 for CPP-Clip. 
    Encoded maps computed by CPP-Clip have the ``red stripe'' when the pitch is around 90$^\circ$ while CPP encoded maps have more smooth transitions when capturing the horizon.
    }
    \vspace{-2mm}
    \label{fig: Encoding noatan}
\end{figure}

Specifically, we set a threshold $\tau$ that represents the prior knowledge of the distance from camera to the furthest point in the scene. Mathematically, for each point $[u, v]$ in the new CPP clipping encoded map $\M_{CPP}^{Clip} \in \mathbb{R^{H \times W}}$, we compute the pseudo depth value:
\begin{equation*}\small
    \M_{CPP}^{Clip}[u, v] = \begin{cases}
          M[u, v] &  M[u, v] < \tau  \\
          \tau & \text{otherwise}
      \end{cases}
\end{equation*}
We set $\tau = 20.0$ in this experiment.
After clipping, we linearly rescale the encoded map to the range of [-1.0, 1.0] to match the statistics of RGB images. We find this yields better performance than directly using $\M_{CPP}^{Clip}$. We visually compare some encoded maps in Fig.~\ref{fig: Encoding noatan}, where we see the clipping method introduces artificial stripes. 
Probably due to this, CPP-Clip does not perform as well as CPP that adopts inverse tangent transform, as shown in Table~\ref{tab: ablation study clipping}.  

\begin{table}[t]
\centering
\caption{\small
Comparisons of different encoding methods evaluated on InteriorNet test-sets. 
CPP applies an inverse tangent transform $\tan^{-1}$ in encoding the camera poses. In contrast, CPP-Clip replaces the $\tan^{-1}$ function with a clipping operation while keeping every other step the same as CPP encoding. 
Both CPP and CPP-Clip perform better than Vanilla model,  demonstrating the effectiveness of our CPP method. Clearly, using the inverse tangent operator is better than clipping. 
}
\begin{adjustbox}{max width=\linewidth}
{
\centering
\begin{tabular}{l| c c | c c }
\hline
\multirow{3}{*}{Models} &  \multicolumn{2}{c|}{\textit{Natural-Test-Set}} & \multicolumn{2}{c}{\textit{Uniform-Test-Set}} \\
& \multicolumn{1}{c|}{\cellcolor{col1} \texttt{$\downarrow$ better}} &  \multicolumn{1}{c|}{\cellcolor[rgb]{0.0,0.8,1.0} \texttt{$\uparrow$ better}}  &
\multicolumn{1}{c|}{\cellcolor{col1} \texttt{$\downarrow$ better}} &  \multicolumn{1}{c}{\cellcolor[rgb]{0.0,0.8,1.0} \texttt{$\uparrow$ better}} \\
& \multicolumn{1}{c|}{\cellcolor{col1}  \small Abs$^r$/Sq$^r$/RMS-log} & \multicolumn{1}{c|}{\cellcolor[rgb]{0.0,0.8,1.0} $\delta^1$ \ / \ $\delta^2$}
& \multicolumn{1}{c|}{\cellcolor{col1} \small Abs$^r$/Sq$^r$/RMS-log} & \multicolumn{1}{c}{\cellcolor[rgb]{0.0,0.8,1.0} $\delta^1$ \ / \ $\delta^2$} \\
\hline
\multicolumn{5}{c}{\cellcolor{lightgrey}\tt InteriorNet} \\
Vanilla
& \multicolumn{1}{c}{.154 / .148 / .229} & \multicolumn{1}{c|}{.803 / .945}
& \multicolumn{1}{c}{.183 / .146 / .250} & \multicolumn{1}{c}{.724 / .926} \\
\hline
+ CPP-Clip
& \multicolumn{1}{c}{.109 / .124 / .204} & \multicolumn{1}{c|}{.871 / .956} 
& \multicolumn{1}{c}{.111 / .096 / .189} & \multicolumn{1}{c}{.867 / .959}\\
+ CPP
& \multicolumn{1}{c}{\bf .108 / .120 / .199} & \multicolumn{1}{c|}{\bf .872 / .958}
& \multicolumn{1}{c}{\bf .106 / .088 / .183} & \multicolumn{1}{c}{\bf .876 / .961} \\
\hline
\multicolumn{5}{c}{\cellcolor{lightgrey}\tt ScanNet} \\
\hline
Vanilla
& \multicolumn{1}{c}{.125 / .068 / .186} & \multicolumn{1}{c|}{.837 / .962}
& \multicolumn{1}{c}{.177 / .121 / .265} & \multicolumn{1}{c}{.711 / .928} \\
\hline
+ CPP-Clip 
& \multicolumn{1}{c}{.110 / .064 / .179} & \multicolumn{1}{c|}{.869 / .964} 
& \multicolumn{1}{c}{.157 / .108 / .248} & \multicolumn{1}{c}{.758 / .936}\\
+ CPP
& \multicolumn{1}{c}{\bf .108 / .060 / .171} & \multicolumn{1}{c|}{\bf .871 / .965}
& \multicolumn{1}{c}{\bf .154 / .106 / .239} & \multicolumn{1}{c}{\bf .781 / .943} \\
\hline
\end{tabular}
}
\end{adjustbox}
\label{tab: ablation study clipping}
\end{table}

\section{Hyperparameter Analysis in CPP Encoding}
\label{ssec:distance_C_study}

\begin{figure}[t]
    \centering
    \includegraphics[width=0.99\linewidth]{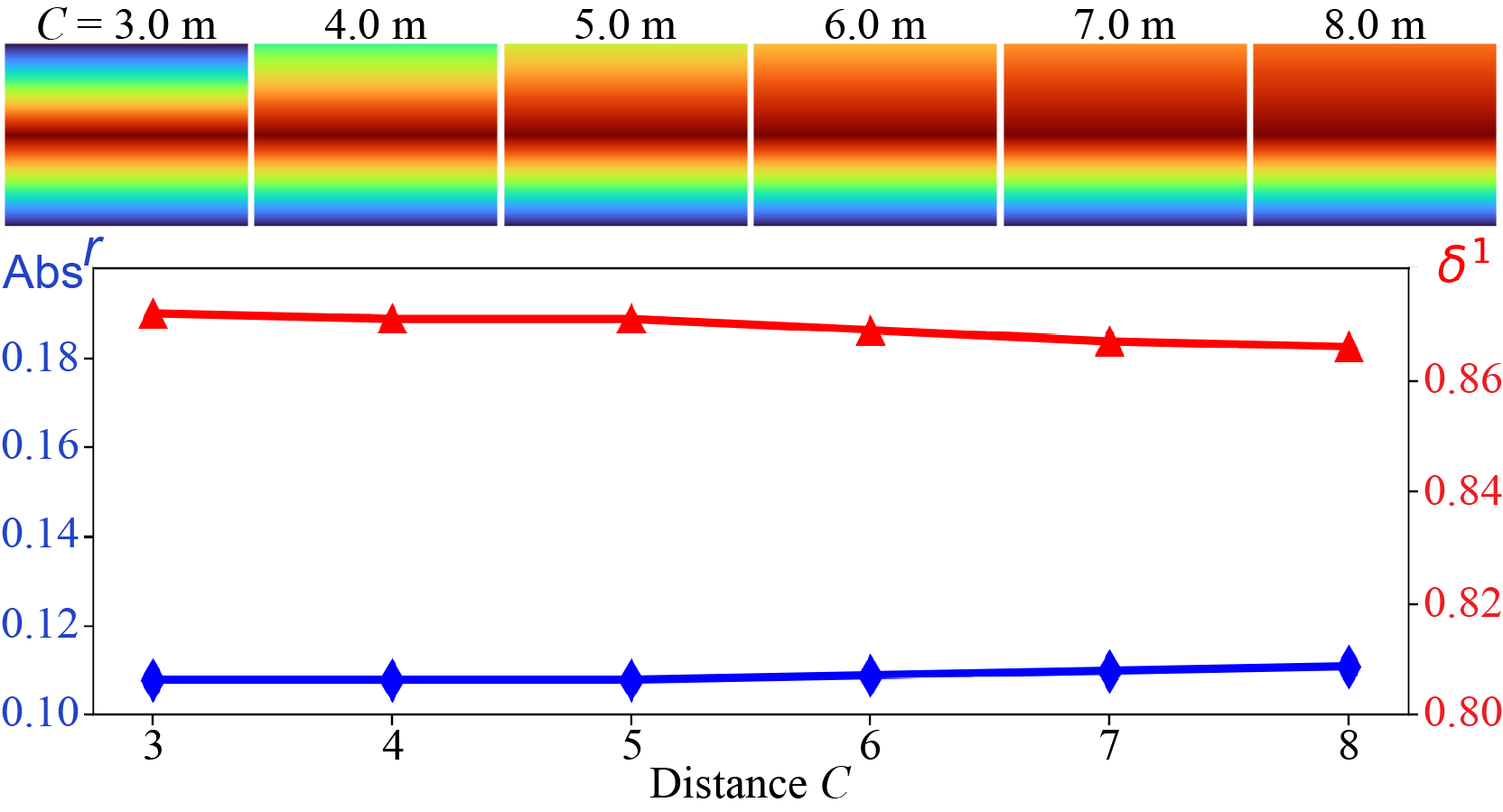}
    \vspace{-3mm}
    \caption{\small
    {\bf Uppeer:} visualizations of CPP encoding with different hyper-parameter $C$ (top); {\bf bottom:} depth prediction performance as a function of hyper-parameter $C$.
    We train depth predictors on InteriorNet \emph{Natural} train-set and test on its \emph{Natural} test-set. From visual inspection, changing the parameter $C$ only affects the part of CPP encoded maps where pixels are above the horizon. As shown by the performance curve, our proposed CPP encoding is very robust w.r.t different values of $C$.
    }
    \vspace{-6mm}
    \label{fig: ceiling_height_study}
\end{figure}

CPP encoding assumes that the camera moves in an empty indoor scene with an infinite floor and ceiling and the distance between two planes in the up direction is described by the parameter $C$.
This distance is set to $C = 3$ meter in all experiments in the main paper. 
To verify the performance change w.r.t the distance $C$, we conduct experiments on InteriorNet \emph{Natural} train-set with various distance $C = [4, 5, 6, 7, 8]$. 
As shown in Fig.~\ref{fig: ceiling_height_study}, the performance of depth predictors are very robust in terms of the parameter $C$. In other words, CPP encoding improves the depth predictor performance and reduces the distribution bias consistently, regardless of the parameter $C$.

\section{Quantitative Results of Blind Predictions}
\label{ssec:blind}
\begin{figure}[t]
\centering
\includegraphics[width=\linewidth]{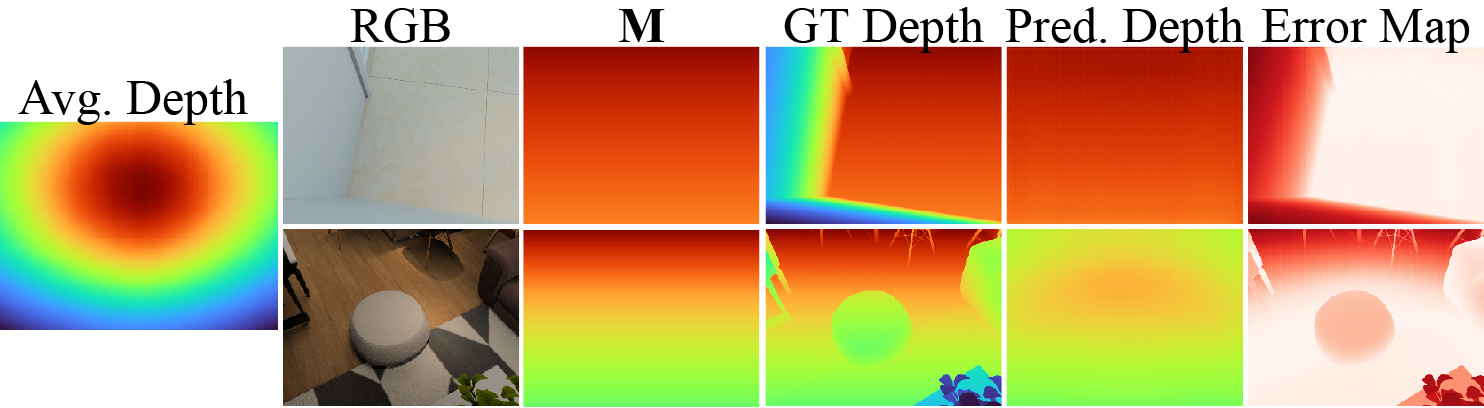}
\vspace{-5mm}
\caption{\small
Illustration of how camera pose provides a strong depth prior through ``blind depth prediction''. Specifically, over the InteriorNet \emph{Natural} train-set, we train a depth predictor solely on the CPP encoded maps $\M$ {\em without} RGB as input. For visual comparison, we compute an averaged depth map (shown left).
We visualize depth predictions on two random examples.
All the depth maps are visualized with the same colormap range.
Perhaps not surprisingly, $\M$ presents nearly the true depth in floor areas, suggesting that camera pose alone does provide strong prior depth information for these scenes.
}
\label{fig: blind prediction2}
\end{figure}

In the main paper, we visually demonstrate that camera poses indeed contain prior knowledge of scene depth. The quantitative results of those visual examples are shown in Table~\ref{tab:blind prediction}, from which we find two key insights. 
First, the blind predictor achieves better performance than evaluating with average training depth maps, suggesting that camera poses alone contain the prior information about scene depth. In other words, training depth predictors with the camera pose alone are better than ``random guess'' from average training depth maps.
Second, the blind predictor achieves promising performance on two images shown in Fig.~\ref{fig: blind prediction2} quantitatively. Together with the visualization of the prediction, we find that blind predictors make significantly more accurate depth prediction on floor regions, which confirms that the camera pose carries the prior knowledge about scene depth, especially on floor and ceiling regions. 

\begin{table}[t]
\setlength\tabcolsep{9pt}
\centering
\captionof{table}{\small
Comparison between using the average depth map (Avg) computed on the InteriorNet \textit{Natural} train-set
and the ``blind predictor'',
which estimates depth solely from per-image CPP encoded maps
\emph{without} RGB images.
We report results on InteriorNet \textit{Natural} test-set.
We find that ``blind predictor'' performs better than ``avg depth map'',
implying the benefit of exploiting camera poses.
We also report on two specific images on which ``blind predictor'' performs well compared to the average performance of Avg or Blind,
as shown in Fig.~\ref{fig: blind prediction2}.
This further confirms that camera poses contain useful prior knowledge about scene depth.
}
\vspace{-2mm}
\begin{adjustbox}{max width=\textwidth}
\small
\begin{tabular}{l| c | c }
\hline
\multirow{2}{*}{Models} &  \multicolumn{1}{c|}{\cellcolor{col1} \texttt{$\downarrow$ better}} &  \multicolumn{1}{c}{\cellcolor[rgb]{0.0,0.8,1.0} \texttt{$\uparrow$ better}} \\
& \cellcolor{col1} { Abs-Rel/Sq-Rel/RMS-log}
& \cellcolor[rgb]{0.0,0.8,1.0}{$\delta^1$ / \ $\delta^2$} \\
\hline
Avg & .414 / .641 / .466 & .346 / .638  \\
Blind & \bf .342 / \bf .519 / \bf .395 & \bf .485 / \bf .750 \\
\hline
Img-1 & .041 / .010 / .082 & .946 / .999  \\
Img-2 & .115 / .059 / .176 & .793 / .990 \\
\hline
\end{tabular}
\end{adjustbox}
\vspace{-3mm}
\label{tab:blind prediction}
\end{table}

\begin{figure*}[t]
\centering
\includegraphics[width=0.99\textwidth]{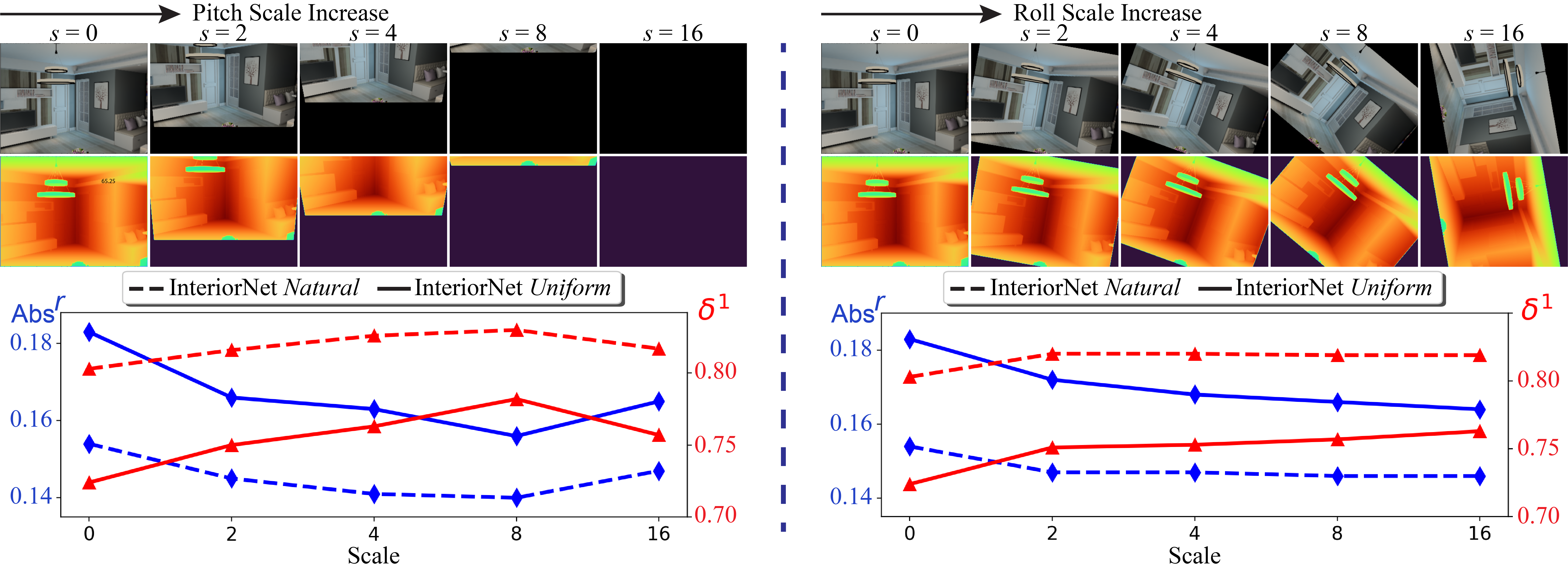}
\vspace{-4mm}
\caption{\small
{\bf Upper row:} visualizations of augmented examples using PDA with different scales. 
{\bf Bottom row:}  performance curves of depth predictor trained with PDA with different scales of pitch $\theta$ (left) and roll $\omega$ (right), respectively.
Please refer to the main paper (Figure 9) for detailed descriptions.
All models are trained on InteriorNet \emph{Natural} train-set and evaluated on both \emph{Natural} (dotted line) and \emph{Uniform} (solid line) test-sets. As we increase the augmentation scale in pitch, the performance of depth predictors improves until scale $s$=16, when large void regions are introduced in the generated examples. On the other hand, increasing augmentation scales in roll lead to steady performance increments. 
In general, PDA consistently improves depth prediction over a Vanilla model trained without PDA.
}
\vspace{-2mm}
\label{fig: aug_study_suppl}
\end{figure*}

\section{Further Study of PDA Augmentation Scales}
\label{ssec:aug}
We provide a more detailed analysis of Vanilla depth predictor plus PDA with different scales of pitch and roll, individually. Please refer to the main paper (Figure 9) for detailed descriptions. As shown in Fig.~\ref{fig: aug_study_suppl}, the performance of the depth predictor monotonically increase until augmenting pitch to the scale of 16. From the visual demonstrations, we believe that the performance drop is due to the introduction of large void regions. On the other hand, by rotating roll, we observe steady performance improvement, which demonstrates that PDA boosts the performance of depth predictors by generating training examples with diverse camera poses. 

\section{Further Study of Camera Height and Rotation in CPP Encoding}
\label{ssec:fixed_pitch_height}
\begin{figure}[h]
\centering
\includegraphics[width=0.95\linewidth]{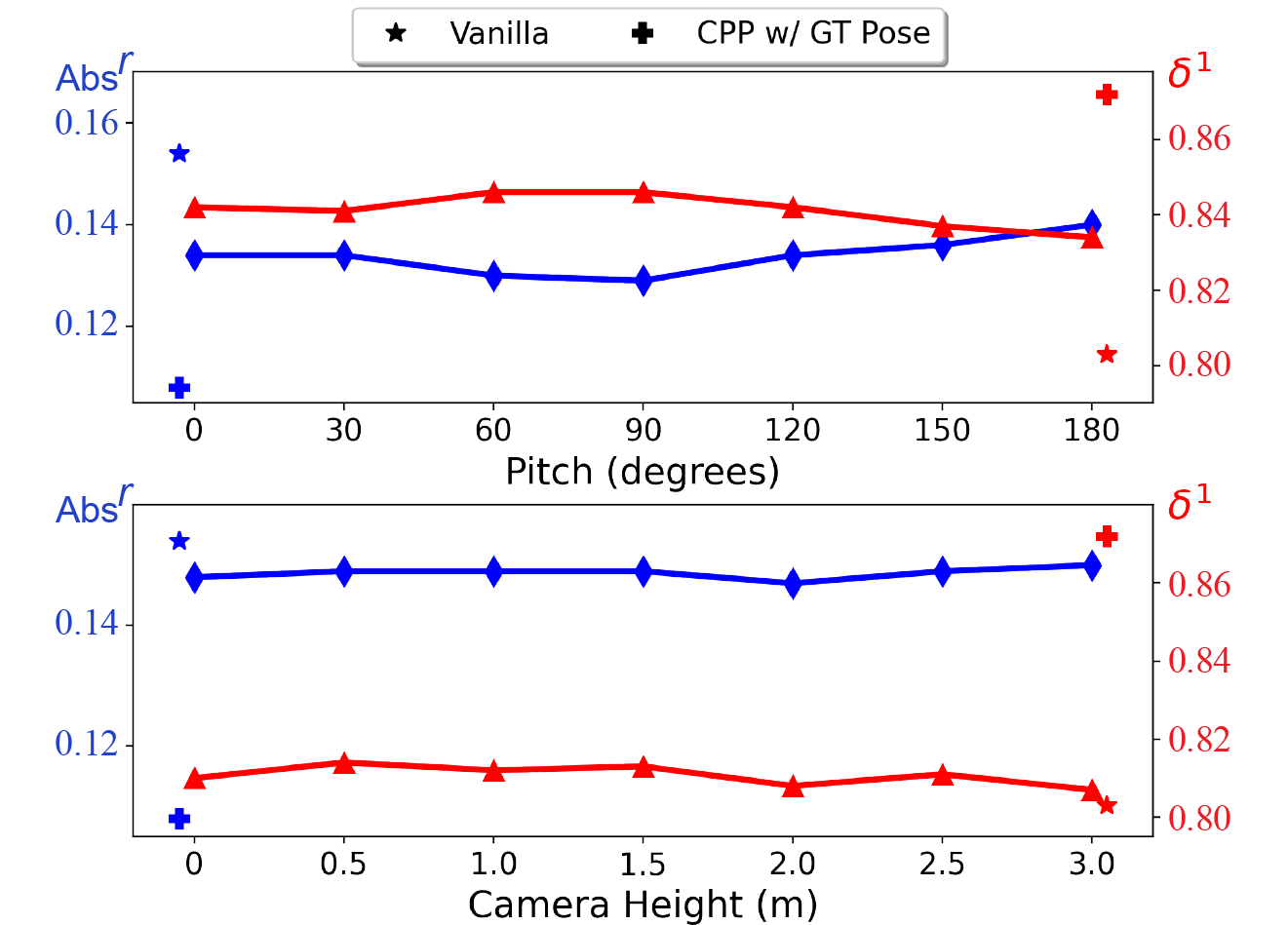}
\vspace{-3mm}
\caption{\small
{\bf Top: }CPP encoding with ground-truth camera height and fixed pitch. {\bf Bottom: } CPP encoding with ground-truth pitch and fixed camera height. All models are trained/evaluated on InteriorNet \emph{Natural} train/test-set. Comparing two blue or red curves across two plots, we find that encoding ground-truth height achieves better performance, suggesting height is ``more important'' than pitch. Moreover, encoding either ground-truth height or pitch outperforms Vanilla model.
}
\label{fig: pitch_camera_height_study}
\end{figure}

CPP encodes rotation (roll and pitch) and camera height, however, 
it is still worth exploring which DOF is more important in CPP encoding.
While it is nontrivial to define ``importance'' as pitch/roll and height have different units and ranges, we did study the pitch and height on InteriorNet (which has a nearly fixed roll). To apply CPP, we fixed either pitch or height and only encode the other with the true value.
As shown in Fig.~\ref{fig: pitch_camera_height_study}, we find that encoding the true camera height (top plot) performs better than the true pitch (bottom plot), and both perform better than the vanilla model. This implies that camera height is ``more important'' than pitch (probably roll as well). 

\section{CPP Encoding with Predicted Poses}
\label{ssec:predicted poses}
We study CPP to encode predicted poses. Specifically, we train depth predictors with CPP using true poses on \emph{Natural} train-sets of the two datasets (Table~\ref{tab:predictive pose}). 
We test models on \emph{Natural} and \emph{Uniform} test-sets, respectively. Note that in testing we encode the predicted poses given by a pose predictor.
As shown in Table~\ref{tab:predictive pose}, CPP with predicted poses still outperforms Vanilla model; when jointly trained with PDA, CPP with predicted poses performs even better.
\begin{table}[t]
\centering
\caption{\small
{\bf CPP Encoding with Predicted Poses}.
We train depth predictors with CPP using true poses on \emph{Natural} train-sets of the two datasets. 
We test models on \emph{Natural} and \emph{Uniform} test-sets, respectively. Note that in testing we encode predicted poses given by a pose predictor.
Clearly, CPP with predicted poses still outperforms Vanilla model; when jointly trained with PDA, CPP with predicted poses performs even better.
Nevertheless, encoding predicted poses underperforms encoding true poses.
}
\vspace{-3mm}
\begin{adjustbox}{max width=\linewidth}
{
\centering
\begin{tabular}{l| c c | c c }
\hline
\multirow{3}{*}{Models} &  \multicolumn{2}{c|}{\textit{Natural-Test-Set}} & \multicolumn{2}{c}{\textit{Uniform-Test-Set}} \\
& \multicolumn{1}{c|}{\cellcolor{col1} \texttt{$\downarrow$ better}} &  \multicolumn{1}{c|}{\cellcolor[rgb]{0.0,0.8,1.0} \texttt{$\uparrow$ better}}  &
\multicolumn{1}{c|}{\cellcolor{col1} \texttt{$\downarrow$ better}} &  \multicolumn{1}{c}{\cellcolor[rgb]{0.0,0.8,1.0} \texttt{$\uparrow$ better}} \\
& \multicolumn{1}{c|}{\cellcolor{col1}  \small Abs$^r$/Sq$^r$/RMS-log} & \multicolumn{1}{c|}{\cellcolor[rgb]{0.0,0.8,1.0} $\delta^1$ \ / \ $\delta^2$}
& \multicolumn{1}{c|}{\cellcolor{col1} \small Abs$^r$/Sq$^r$/RMS-log} & \multicolumn{1}{c}{\cellcolor[rgb]{0.0,0.8,1.0} $\delta^1$ \ / \ $\delta^2$} \\
\hline
\multicolumn{5}{c}{\cellcolor{lightgrey}\tt InteriorNet} \\
Vanilla
& \multicolumn{1}{c}{.154 / .148 / .229} & \multicolumn{1}{c|}{.803 / .945}
& \multicolumn{1}{c}{.183 / .146 / .250} & \multicolumn{1}{c}{.724 / .926} \\
\hline
+ CPP$_{pred}$
& \multicolumn{1}{c}{.142 / .132 / .212} & \multicolumn{1}{c}{.825 / .951}
& \multicolumn{1}{c}{.164 / .121 / .228} & \multicolumn{1}{c}{.756 / .946} \\
+ CPP
& \multicolumn{1}{c}{.108 / .120 / .199} & \multicolumn{1}{c|}{.872 / .958}
& \multicolumn{1}{c}{.106 / .088 / .183} & \multicolumn{1}{c}{.876 / .961} \\
\hline
+ Both$_{pred}$
& \multicolumn{1}{c}{.135 / .127 / .205} & \multicolumn{1}{c|}{.849 / .955}
& \multicolumn{1}{c}{.148 / .114 / .213} & \multicolumn{1}{c}{.780 / .952} \\
+ Both
& \multicolumn{1}{c}{\bf .095 / .101 / .180} & \multicolumn{1}{c|}{\bf .898 / .966}
& \multicolumn{1}{c}{\bf .091 / .069 / .161} & \multicolumn{1}{c}{\bf .903 / .973} \\
\hline
\multicolumn{5}{c}{\cellcolor{lightgrey}\tt ScanNet} \\
Vanilla
& \multicolumn{1}{c}{.125 / .068 / .186} & \multicolumn{1}{c|}{.837 / .962}
& \multicolumn{1}{c}{.177 / .121 / .265} & \multicolumn{1}{c}{.711 / .928} \\
\hline
+ CPP$_{pred}$
& \multicolumn{1}{c}{.116 / .065 / .180} & \multicolumn{1}{c|}{.852 / .964}
& \multicolumn{1}{c}{.169 / .117 / .255} & \multicolumn{1}{c}{.731 / .931} \\
+ CPP
& \multicolumn{1}{c}{.108 / .060 / .171} & \multicolumn{1}{c|}{.871 / .965}
& \multicolumn{1}{c}{.154 / .106 / .239} & \multicolumn{1}{c}{.781 / .943} \\
\hline
+ Both$_{pred}$
& \multicolumn{1}{c}{.111 / .061 / .173} & \multicolumn{1}{c|}{.866 / .965}
& \multicolumn{1}{c}{.159 / .111 / .247} & \multicolumn{1}{c}{.773 / .938} \\
+ Both
& \multicolumn{1}{c}{\bf .102 / .052 / .160} & \multicolumn{1}{c|}{\bf .882 / .973}
& \multicolumn{1}{c}{\bf .143 / .097 / .230} & \multicolumn{1}{c}{\bf .809 / .952} \\
\hline
\end{tabular}
}
\end{adjustbox}
\vspace{-4mm}
\label{tab:predictive pose}
\end{table}

\begin{figure}[h]
\centering
\includegraphics[width=0.95\linewidth]{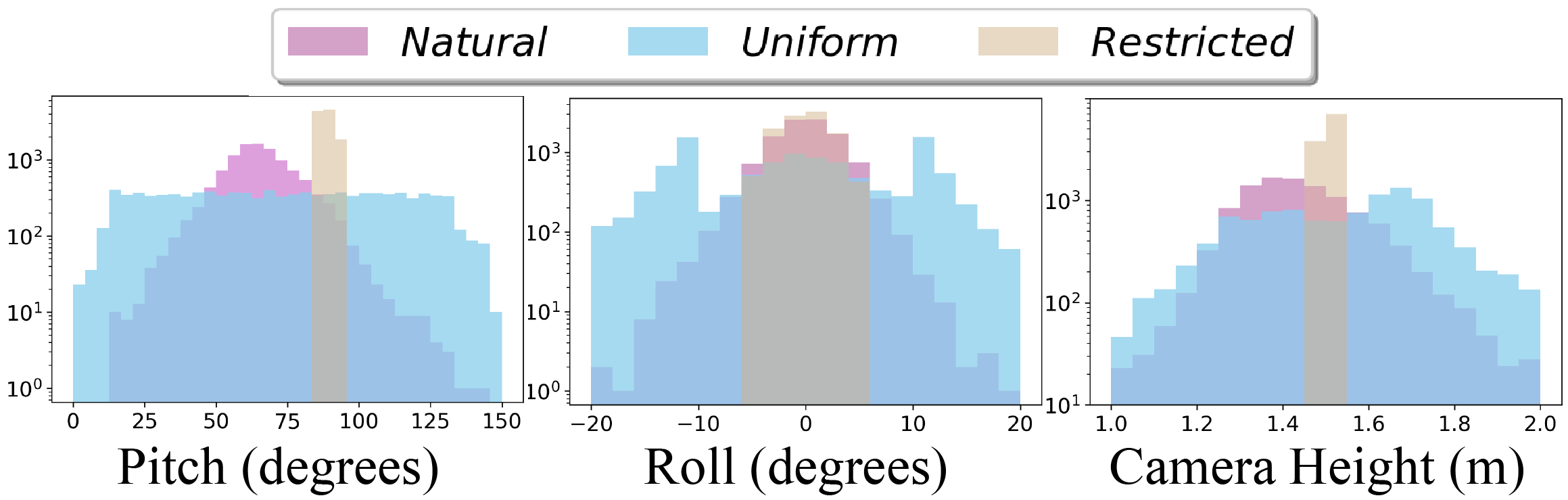}
\vspace{-3mm}
\caption{\small
Distribution of pitch, roll and camera height for three subsets of images from ScanNet.  
From the \emph{Natural} subset, we observe the ScanNet dataset also has a naturally biased distribution in both pitch, roll and camera height.
Please refer to Section 5 in the main paper on how we construct these three subsets.
}
\vspace{-4mm}
\label{fig: ScanNet-distribution-figure}
\end{figure}

\section{Additional Details in Experiments}
\label{ssec:impl}

\subsection{Image and Depth Preprocessing}
All input RGB images are first normalized to the range of $[-1.0, 1.0]$ and then resized to $240 \times 320$ before feeding into CNNs. 
Note that resizing images to $240 \times 320$ does not change their original aspect ratios. 
For better training, as a preprocessing step on the depth~\cite{bian2019unsupervised, godard2017unsupervised},
we apply the following operation to rescale depth maps $y$ to get a normalized map $y'$:
\begin{equation}
    y' = (\frac{y - E_{min}}{E_{max} - E_{min}} - 0.5) * 2.0,
\label{eq:pre-processing}
\end{equation}
where $E_{min} = 1.0$ and $E_{max} = 10.0$ are the minimum and maximum evaluation values, respective.
The above operation is a map from $[1.0, 10.0]$ to $[-1.0, 1.0]$.
In the literature, it is reported the model can be trained better in this scale range~\cite{zheng2018t2net, zhao2019geometry}.
We only compute the loss for pixels that have depth values between 1.0 and 10.0 meters. 
We evaluate the depth prediction on the original depth scale. To do so, we apply an inverse operation of Eq.~\ref{eq:pre-processing} to the predicted depth maps. Moreover, we also only evaluate the depth that lies in [1, 10] meters.

\subsection{Pose Prediction Network}
When camera poses are not available during testing, we train a camera pose predictor that predicts camera pitch $\theta$, roll $\omega$ and height $h$ for CPP encoding (i.e., the CPP$_{pred}$ model). We build the pose predictor over ResNet18 structure with a new top layer that outputs a 3-dim vector to regress pitch, roll, and camera height.
During training, we load the ImageNet pretrained weights and finetune the weights for pose predictions with L1 loss.



\subsection{Evaluation Protocol}
The depth evaluation range in this work is from 1.0m to 10.0m for both InteriorNet and ScanNet.
For each method, we save a checkpoint every 10 epochs and select the checkpoint that produces the smallest average L1 loss on the validation set to report the performance.



\subsection{ScanNet Camera Pose Distribution}
The camera pose distribution of subsets in ScanNet is shown in Fig.~\ref{fig: ScanNet-distribution-figure}.
While it is hard to sample a subset with exactly uniform distribution w.r.t to all attributes (i.e., pitch, roll, and camera height), we sample the \emph{Uniform} subset with the priority of pitch, roll, height from high to low.
As these subsets differ a lot in terms of camera pose distribution, they serve our study w.r.t camera distribution bias.

\begin{figure*}[t]
    \centering
    \includegraphics[width=0.9\textwidth]{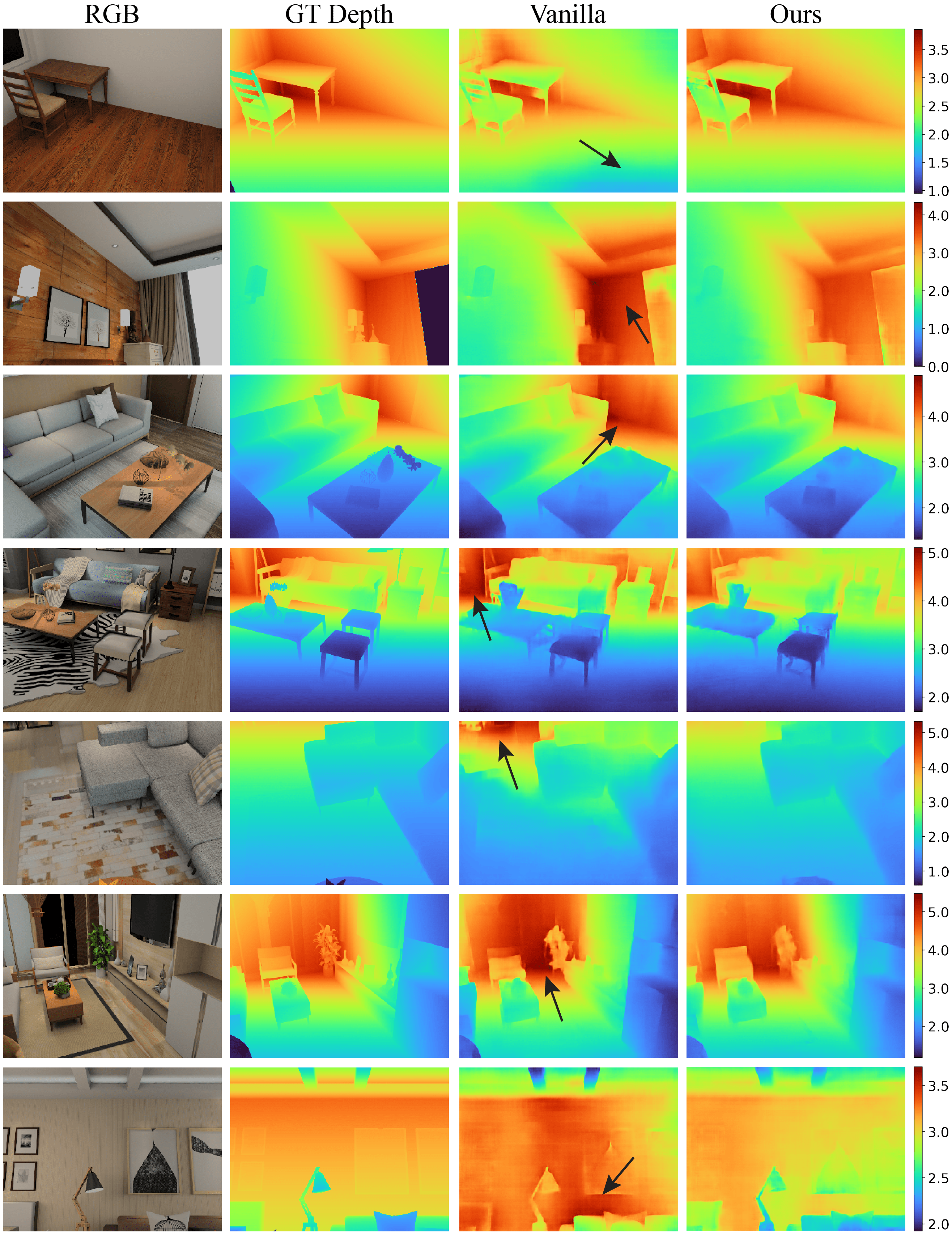}
    \caption{
    Depth predictions of Vanilla and our model (jointly applied CPP and PDA) on InteriorNet test-set.
    From these images captured under various camera poses, our model predicts better depth than Vanilla model in terms of the overall scale.
    }
    \label{fig:results-1}
\end{figure*}

\begin{figure*}[t]
    \centering
    \includegraphics[width=0.9\textwidth]{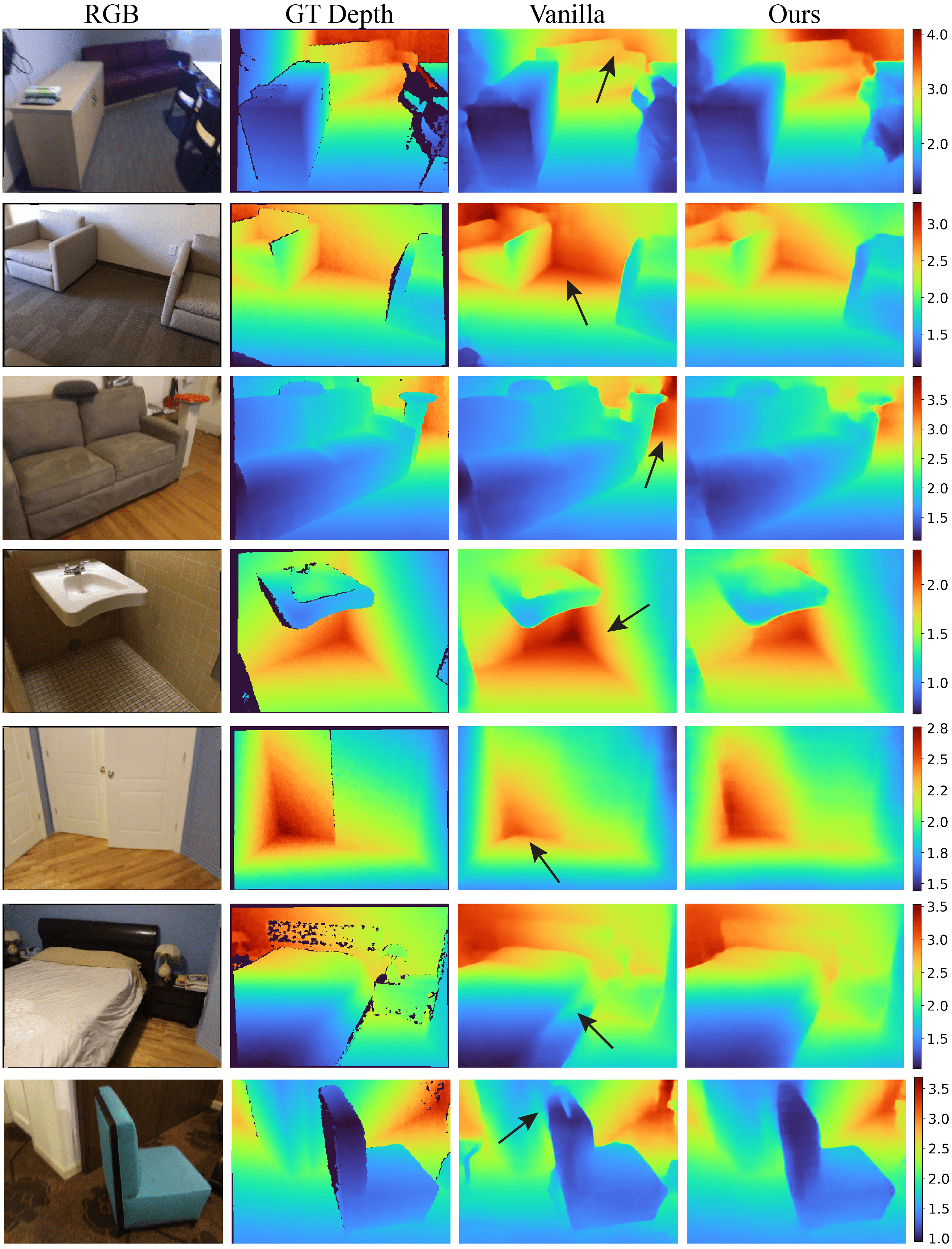}
    \caption{
    Depth predictions of Vanilla and our model (jointly applied CPP and PDA) on ScanNet test-set.
    From these images captured under various camera poses, our model predicts better depth than Vanilla model in terms of the overall scale.
    }
    \label{fig:results-2}
\end{figure*}

\end{document}